\documentclass[11pt]{article} 

\usepackage{hyperref}[links]
\hypersetup{
  final,
  citecolor = darkgreen,
    colorlinks=true,
    linkcolor=blue,
    filecolor=magenta,      
    urlcolor=cyan
}

\usepackage{tcolorbox}

\usepackage{url}
\usepackage{mathtools}
\usepackage{fullpage}

\usepackage{amsmath,amssymb,amsthm, amsfonts}
\usepackage{mathbbol}
\usepackage{mathrsfs}  
\usepackage{multicol}
\usepackage{rotating}
\usepackage{enumitem, comment, xifthen}

\usepackage{xcolor}
\definecolor{darkgreen}{rgb}{0.0, 0.5, 0.0}

\usepackage{mathbbol}

\usepackage{booktabs}

\usepackage[nameinlink]{cleveref}

\usepackage[mathscr]{euscript}

\newcommand{\bcar}{\begin{itemize}}
\newcommand{\ecar}{\end{itemize}}

\newcommand{\SUMN}{\ensuremath{\sum_{i=1}^\numobs}}

\newcommand{\Prob}{\ensuremath{\mprob}}

\newcommand{\Rhat}{\ensuremath{\widehat{\Risk}_\numobs}}
\newcommand{\Risk}{\ensuremath{R}}

\newcommand{\nucnorm}[1]{\ensuremath{\matsnorm{#1}{\mbox{\tiny{nuc}}}}}
\newcommand{\frobnorm}[1]{\ensuremath{\matsnorm{#1}{\mbox{\tiny{F}}}}}

\definecolor{MyGray}{rgb}{0.9,0.9,0.9}

\makeatletter\newenvironment{graybox}{ 
\begin{lrbox}{\@tempboxa}
\begin{minipage}{0.985\columnwidth}}{\end{minipage}
\end{lrbox}%
\colorbox{MyGray}{\usebox{\@tempboxa}} }
\makeatother



\newcommand{\Zback}[1]{\ensuremath{Z^{\backslash i}}}

\newcommand{\xsamstack}[1]{\ensuremath{x_1^\numobs}}
\newcommand{\Xsamstack}[1]{\ensuremath{X_1^\numobs}}

\newcommand{\widgraph}[2]{\includegraphics[keepaspectratio,width=#1]{#2}}

\newcommand{\Term}{\ensuremath{T}}

\newcommand{\radepl}{\ensuremath{\varepsilon}}
\newcommand{\rade}[1]{\ensuremath{\radepl_{#1}}}

\newcommand{\Event}{\ensuremath{\mathcal{E}}}

\newcommand{\Fclass}{\ensuremath{\mathscr{F}}}

\newcommand{\fancysoln}[1]{
\ifthenelse{\equal{\doctype}{WITHSOLS}}
{
\begin{soln}
#1
\end{soln}
}
{
}
}

\newcommand{\wtil}{\ensuremath{\widetilde{w}}}

\newcommand{\ftil}{\ensuremath{\widetilde{f}}}

\newcommand{\fstar}{\ensuremath{f^*}}

\long\def\comment#1{}

\makeatletter
\def\@cite#1#2{[\if@tempswa #2 \fi #1]}
\makeatother

\newcommand{\fhat}{\ensuremath{\widehat{f}}}
\newcommand{\yhat}{\ensuremath{\widehat{y}}}

\newcommand{\Xspace}{\ensuremath{\mathcal{X}}}

\newcommand{\defn}{\vcentcolon=}

\newcommand{\real}{\ensuremath{\mathbb{R}}}

\newcommand{\numobs}{\ensuremath{n}}

\newcommand{\mprob}{\ensuremath{\mathbb{P}}}

\newcommand{\mymathbf}[1]{\ensuremath{\mathbf{#1}}}

\newcommand{\Ymat}{\ensuremath{\mymathbf{Y}}}

\newcommand{\Smat}{\ensuremath{\mymathbf{S}}}
\newcommand{\Wmat}{\ensuremath{\mymathbf{W}}}

\newcommand{\Rmat}{\ensuremath{\mymathbf{R}}}

\newlength{\widebarargwidth}
\newlength{\widebarargheight}
\newlength{\widebarargdepth}
\DeclareRobustCommand{\widebar}[1]{%
  \settowidth{\widebarargwidth}{\ensuremath{#1}}%
  \settoheight{\widebarargheight}{\ensuremath{#1}}%
  \settodepth{\widebarargdepth}{\ensuremath{#1}}%
  \addtolength{\widebarargwidth}{-0.3\widebarargheight}%
  \addtolength{\widebarargwidth}{-0.3\widebarargdepth}%
  \makebox[0pt][l]{\hspace{0.3\widebarargheight}%
    \hspace{0.3\widebarargdepth}%
    \addtolength{\widebarargheight}{0.3ex}%
    \rule[\widebarargheight]{0.95\widebarargwidth}{0.1ex}}%
  {#1}}

\newcommand{\matsnorm}[2]{|\!|\!| #1 |\!|\!|_{{#2}}}

\newcommand{\inprod}[2]{\ensuremath{\langle #1 , \, #2 \rangle}}

\newcommand{\Exs}{\ensuremath{\mathbb{E}}}

\usepackage{mynew_macros}

\newcommand{\noise}{\ensuremath{w}}

\newcommand{\FULLSUM}{\ensuremath{\frac{1}{\numobs} \SUMN}}

\newcommand{\revdefn}{\ensuremath{= \colon}}

\newcommand*\mybar[1]{%
  \: \,
  \hbox{%
    \vbox{%
      \hrule height 0.75pt 
      \kern0.4ex
      \hbox{%
        \kern-0.1em
        \ensuremath{#1}%
        \kern-0.1em
      }%
    }%
  }%
}

\newcommand{\Zopt}{\ensuremath{Z_{\numobs}}}
\newcommand{\Ztrue}{\ensuremath{\mybar{Z}_\numobs}}

\newcommand{\Zwild}{\ensuremath{W_\numobs}}
\newcommand{\ZwildRade}{\ensuremath{\tinysuper{W}{ideal}_\numobs}}

\newcommand{\oneplus}{\ensuremath{1 + \tfrac{1}{s}}}
\newcommand{\twoplus}{\ensuremath{2 + \tfrac{1}{s}}}

\newcommand{\MyBall}[2]{\ensuremath{\mathbb{B}_{#2}(#1)}}

\newcommand{\ZnewHat}{\ensuremath{Z}_\numobs}

\newcommand{\fproj}{\ensuremath{f^\dagger}}

\theoremstyle{plain}
\newtheorem{theorem}{Theorem}
\newtheorem{lemma}{Lemma}

\theoremstyle{definition}

\theoremstyle{remark}

\makeatletter
\long\def\@makecaption#1#2{
        \vskip 0.8ex
        \setbox\@tempboxa\hbox{\small {\bf #1:} #2}
        \parindent 1.5em  
        \dimen0=\hsize
        \advance\dimen0 by -3em
        \ifdim \wd\@tempboxa >\dimen0
                \hbox to \hsize{
                        \parindent 0em
                        \hfil 
                        \parbox{\dimen0}{\def\baselinestretch{0.96}\small
                                {\bf #1.} #2
                                } 
                        \hfil}
        \else \hbox to \hsize{\hfil \box\@tempboxa \hfil}
        \fi
        }
\makeatother

\newcommand{\Exrisk}{\ensuremath{\mathcal{E}}}

\newcommand{\ExHat}{\ensuremath{\widehat{\Exrisk}_\numobs}}
\newcommand{\ExPop}{\ensuremath{\widebar{\Exrisk}}}

\newcommand{\rhat}{\ensuremath{\widehat{r}_\numobs}}

\newcommand{\empinner}[2]{\ensuremath{\inprod{#1}{#2}_\numobs}}
\newcommand{\empnorm}[1]{\ensuremath{\|#1\|_\numobs}}
\newcommand{\empnormsq}[1]{\ensuremath{\|#1\|^2_\numobs}}

\newcommand{\noisetil}{\ensuremath{\widetilde{w}}}

\newcommand{\figdir}{.}

\usepackage{scalerel}

\newcommand{\tinysuper}[2]{#1^{\scaleto{\mbox{#2}}{3.5pt}}}

\newcommand{\wnoise}{\ensuremath{w}}

\newcommand{\nstd}{\ensuremath{\rho}}

\newcommand{\wild}[1]{\tinysuper{#1}{$\blacklozenge$}}

\newcommand{\ywild}{\ensuremath{\wild{y}}}
\newcommand{\wwild}{\ensuremath{\wild{w}}}
\newcommand{\wwildrho}{\ensuremath{\wild{w}_\nstd}}

\newcommand{\fwildrho}{\ensuremath{\wild{f}_\nstd}}
\newcommand{\fwild}{\fwildrho}

\newcommand{\rwild}{\ensuremath{\wild{r}_\nstd}}

\newcommand{\MyPlainOpt}{\ensuremath{\mathrm{Opt}}}
\newcommand{\Opt}{\ensuremath{\MyPlainOpt^\star}}
\newcommand{\OptWild}{\ensuremath{\wild{\widetilde{\MyPlainOpt}}}}
\newcommand{\Meth}{\ensuremath{\mathcal{M}}}

\newcommand{\Pen}{\ensuremath{\mathcal{P}}}

\newcommand{\DevTerm}{\ensuremath{\tfrac{2 \|\wnoise\|_\infty
        \tpar}{\sqrt{\numobs}}}}

\newcommand{\ZoptRade}{\ensuremath{Z_\numobs^{\rade{}}}}

\newcommand{\ExsRade}{\ensuremath{\Exs_{\rade{}}}}

\newcommand{\radetil}{\ensuremath{\tilde{\rade{}}}}

\newcommand{\tpar}{\ensuremath{t}}

\newcommand{\AppError}{\ensuremath{A_\numobs(\ftil)}}

\renewcommand{\defn}{\ensuremath{: \, =}}

\newcommand{\HighPlain}{\ensuremath{H}}

\newcommand{\twoplust}{\ensuremath{2 + \tfrac{1}{t}}}

\newcommand{\HackHigh}{\ensuremath{\tfrac{6 \|\noise\|_\infty}{t}
    \rhat^2}}

\newcommand{\resp}{\ensuremath{y}}
\newcommand{\covariate}{\ensuremath{x}}

\newcommand{\OptDagger}{\ensuremath{\MyPlainOpt^\dagger}}

\newcommand{\Convex}{\ensuremath{\mathcal{C}}}

\newcommand{\SmatStar}{\ensuremath{\mathbf{S^*}}}
\newcommand{\numframes}{m}
\newcommand{\numpoints}{p}
\newcommand{\numpoint}{\numpoints}

\newcommand{\gamwild}{\ensuremath{\wild{\gamma}_\rho}}
\newcommand{\ShatEst}{\ensuremath{\widehat{\Smat}}}

\newcommand{\Rbarpl}{\ensuremath{\bar{\mathcal{R}}}}
\newcommand{\Rbar}{\ensuremath{\widebar{\mathcal{R}}}}

\begin{document}

\begin{center}
  {\bf{\LARGE  Wild refitting for black box prediction}}

    \vspace*{.2in}
  \begin{tabular}{c}
    Martin J. Wainwright
  \end{tabular}
  
  \begin{tabular}{c}
   Laboratory for Information and Decision Systems
   \\
   Statistics and Data Science Center \\
   EECS and Mathematics \\
   Massachusetts Institute of Technology
  \end{tabular}

  \vspace*{0.2in}

    July 8, 2025
  
  \vspace*{0.2in}

  \begin{abstract}
    We describe and analyze a computationally efficient refitting
    procedure for computing high-probability upper bounds on the
    instance-wise mean-squared prediction error of penalized
    nonparametric estimates based on least-squares
    minimization. Requiring only a single dataset and black box access
    to the prediction method, it consists of three steps: computing
    suitable residuals, symmetrizing and scaling them with a
    pre-factor $\rho$, and using them to define and solve a modified
    prediction problem recentered at the current estimate.  We refer
    to it as wild refitting, since it uses Rademacher residual
    symmetrization as in a wild bootstrap variant.  Under relatively
    mild conditions allowing for noise heterogeneity, we establish a
    high probability guarantee on its performance, showing that the
    wild refit with a suitably chosen wild noise scale $\rho$ gives an
    upper bound on prediction error. This theoretical analysis
    provides guidance into the design of such procedures, including
    how the residuals should be formed, the amount of noise rescaling
    in the wild sub-problem needed for upper bounds, and the local
    stability properties of the block-box procedure.  We illustrate
    the applicability of this procedure to various problems, including
    non-rigid structure-from-motion recovery with structured matrix
    penalties; plug-and-play image restoration with deep neural
    network priors; and randomized sketching with kernel methods.
  \end{abstract}

\end{center}


\section{Introduction}

Prediction problems are ubiquitous throughout science and engineering.
They are defined by a feature vector or covariate $X \in \Xspace$ and
a scalar response $Y \in \real$, and the goal is to determine a
function $x \mapsto \fhat(x)$ that is a ``good approximation'' to $y$
in a certain sense.  When the quality of an estimate is measured by
its mean-squared error, then it is a classical fact that the best
predictor is given by the regression function $\fstar(x) = \Exs[Y \mid
  X= x]$.  The quality of a given estimate $\fhat$ can then be
measured by its mean-squared error or risk $\Rbar(\fhat) \defn
\Exs_{X,Y} \big[ (Y- \fhat(X))^2$, or its excess risk
\begin{align}
\label{EqnDefnPopExcRisk}  
  \ExPop(\fhat) & \defn \underbrace{\Exs_{X,Y} \big[ (Y- \fhat(X))^2
      \Big]}_{\Rbarpl(\fhat)} - \Exs_{X, Y} \big[ (Y- \fstar(X))^2
    \big].
\end{align}
Here the expectations are taken over a future sample $(X, Y)$,
independent of any training data used to produce $\fhat$.  Since
$\fstar$ is the optimal predictor in the mean-squared sense, the
excess risk is always non-negative.

\subsection{Methods for risk estimation}

Given a predictor $\fhat$ that has been fit on some training data, an
important problem in practice is to estimate the excess risk
$\ExPop(\fhat)$, or simply the risk $\Rbar(\fhat)$ itself.  Given its
importance, this problem has been intensively studied, with a variety
of approaches available.  The simplest and most classical approach is
the \emph{hold-out method}; we split the original dataset into two
separate pieces, and use the first piece to fit the estimate $\fhat$,
and the second piece to obtain an unbiased estimate of the risk
$\Rbar(\fhat)$.  However, the hold-out approach suffers from a number
of deficiencies, notably its wastefulness in data use.  A more
sophisticated approach, dating back to the classic
papers~\cite{stone1974cross,geisser1975predictive}, is
cross-validation; see the paper~\cite{arlot2010survey} for a modern
overview.  Cross-validation involves repeated splitting of the dataset
into parts used for training and testing; in this way, it avoids the
wasteful use of data associated with the hold-out method.  On the flip
side, it is significantly more computationally expensive; for example,
roughly speaking, a CV approach based on $k$-splits will be require
$k$-times more computation than hold-out.  Note that hold-out and
cross-validation apply to i.i.d.  datasets\footnote{There are
relaxations of the i.i.d. condition, but the essential requirement is
that the risks across different splits of the dataset are
equivalent.}, so that risks are preserved in expectation across
splits.  It is also possible to perform risk estimation via various
types of resampling methods, including both parametric and
non-parametric forms of the
bootstrap~\cite{efron2004estimation,efron1993introduction}.  Of
particular relevance to this paper is the \emph{wild bootstrap}
(e.g.,~\cite{wu1986jackknife,liu1988bootstrap,li1989resampling,mammen1993bootstrap}),
a bootstrapping procedure that is well-suited to heteroskedastic data;
we discuss it at more length below.  For problems with particular
structure, other more specialized methods are available.  Notably, if
one is willing to assume that the effective noise $Y - \fstar(X)$
follows a Gaussian distribution, then Stein's unbiased estimate of
risk~\cite{stein1981estimation}, along with the associated family of
covariance-based penalties~\cite{efron2004estimation}, be used for
risk estimation.  It is also possible to combine bootstraps with
covariance-based penalties~\cite{tibshirani2009bias}.

It should be noted that many methods---among them cross-validation and
various types of covariance-based penalties---do \emph{not} provide
estimates of the risk $\Rbar(\fhat)$; this risk is itself is a random
variable, due to the data used to train $\fhat$. Instead, such methods
are targeting the deterministic quantity obtained by averaging
$\Rbar(\fhat)$ over the training data.  Bates et
al.~\cite{bates2024cross} provide discussion of this distinction and
related subtleties in the context of cross-validation.  To make this
distinction clear, we refer to the random variable $\Rhat(\fhat)$---or
the corresponding excess risk $\ExPop(\fhat)$---as the
\emph{instance-wise risk.}

\subsection{Contributions}
\label{SecContributions}
This paper is motivated by the goal of obtaining a method for risk
estimation with the following properties:
\begin{enumerate}
\item[(a)] For a given predictor $\fhat$, it provides an estimate of
  the instance-wise risk $\Rhat(\fhat)$ (or the instance-wise excess
  risk $\ExPop(\fhat)$), and this estimate is equipped with an
  explicit and non-asymptotic guarantee.
\item[(b)] It allows for heteroskedasticity, and other forms of
  heterogeneity in the distribution of pairs $(x_i, y_i)$ used to fit
  the predictor $\fhat$.
\item[(c)] It requires only black-box access to the statistical method
  $\Meth$ for producing predictors---that is, we can re-fit the method
  for a new dataset, but we have no visibility into its inner
  structure.
\item[(d)] It makes a limited number of queries to the method $\Meth$.
\end{enumerate}

Let us provide some motivation for these desiderata.  Beginning with
requirement (a), it is desirable to target the instance-wise risk
$\Rbar(\fhat)$, since it correctly reflects the mean-squared error
that will achieved by applying the given predictor $\fhat$---with the
underlying training data held fixed---on future data.  Non-asymptotic
bounds are desirable in that they are explicit and hold for all sample
sizes (as opposed to an asymptotic statement).  Re requirement (b),
modern datasets often exhibit various forms of heterogeneity, so it is
desirable to weaken independence or exchangeability assumptions as
much as possible.  As for the black-box requirement in item (c), many
methods in modern statistics and machine learning are relatively
complex, and it is desirable to have methods for risk estimation that
do not require explicit knowledge about their structure.  Deep neural
networks, which are widely used for prediction, are a canonical
instance of such an opaque model.  Lastly, the limited query condition
is item (d) is also important in modern statistical practice;  the
training process required to fit a method $\Meth$ can be quite expensive,
so it is undesirable to have a procedure (e.g., such as a typical
bootstrap) that requires repeated re-fitting of the model a large number
of times.

In this paper, we describe a procedure for risk estimation that is
equipped with properties (a) through (d).  It is inspired by a classic
line of work on the wild bootstrap
(e.g.,~\cite{wu1986jackknife,liu1988bootstrap,li1989resampling,
  mammen1992bootstrap}), and for this reason, we refer to it as the
\emph{wild refitting procedure}.  At its core is a form of Rademacher
symmetrization applied to a set of estimated residuals;
see~\Cref{SecWildRefit} for a detailed description.  This type of
Rademacher symmetrization is standard in the analysis of uniform laws
of large numbers and related empirical
processes~\cite{vanderVaart96,vandeGeer,Wai19}. In the context of the
wild bootstrap, it has also been proposed and
studied~\cite{liu1988bootstrap,mammen1992bootstrap,mammen1993bootstrap},
with theoretical results given on its asymptotic behavior.  In
contrast, we analyze this form of Rademacher in a non-asymptotic
setting, making essential use of the sharp concentration properties of
Lipschitz functions of Rademacher variables~\cite{Ledoux01}.

\paragraph{Organization:}
The remainder of this paper is organized in follows.  We begin
in~\Cref{SecFix} by setting up the problem, whereas~\Cref{SecExcess}
is devoted to the notion of \emph{optimism} that distinguishes between
the population and empirical risks.  In~\Cref{SecWildRefit}, we
describe the wild refitting procedure that is the main focus of this
paper, and we provide a simple illustration of its behavior
in~\Cref{SecTik}.  Our main theoretical results are given
in~\Cref{SecMain}, including~\Cref{ThmWildOptBound} that bounds the
excess risk in terms of a quantity known as the wild optimism,
and~\Cref{ThmRhatError} that shows how to bound the statistical
estimation error.  In~\Cref{SecNumerical}, we report the results of
some numerical studies, where we illustrate the use of wild refitting
for non-rigid structure-from-motion recovery (\Cref{SecDance}) and
image denoising methods using plug-and-play methods with deep neural
net priors (\Cref{SecPlug}).  We provide proofs in~\Cref{SecProofs},
and conclude with a discussion in~\Cref{SecDiscuss}.


\section{Problem set-up}
\label{SecSetup}

In this section, we begin by setting up the problem more precisely,
before describing the wild refitting procedure that we study in this paper.

\subsection{$M$-estimation and fixed design prediction}
\label{SecFix}

Given a covariate-response pair $(X, Y)$, the optimal function
$\fstar$ for generating predictions $x \mapsto \yhat$ is given by the
\emph{regression function} \mbox{$\fstar(x) \defn \Exs[Y \mid X =
    x]$.}  Thus, given a collection of $\numobs$ covariate-response
pairs $(x_i, y_i)$, they can viewed as generated from the model
\begin{align*}
y_i & = \fstar(x_i) + w_i \qquad \mbox{for $i = 1, \ldots, \numobs$}
\end{align*}
where each $w_i$ is a realization of a conditionally zero-mean random
variable $W_i$ (i.e., such that $\Exs[W_i \mid X_i = x_i] = 0$.)  The
focus of this paper is the \emph{fixed design} case, where we
condition upon the given realization $\{x_i \}_{i=1}^\numobs$ of the
covariates, and compute prediction errors using the empirical
distribution $\Prob_n$ that places mass $1/\numobs$ at each $x_i$.

On one hand, this type of conditional analysis can be performed for
any prediction problem; moreover, there are many prediction problems
where it is most natural to view the covariates as fixed.  Examples
include prediction of a time series, where covariates correspond to
particular points in time; and prediction in images or video, where
covariates correspond to particular two-dimensional spatial positions
(for an image), or 2D-space-time positions (for a video).

\paragraph{Prediction via $M$-estimation:}
In this paper, we study prediction methods that are specified by a
real-valued class $\Fclass$ of functions on the covariate space
$\Xspace$, and a penalty function $\Pen: \Fclass \rightarrow \real$.
Given a data set $\{(x_i, y_i) \}_{i=1}^\numobs$, we obtain the fitted
predictor $\fhat$ by solving the optimization problem
\begin{align}
\label{EqnMest}
\fhat \in \arg \min_{f \in \Fclass} \Big \{ \FULLSUM \big(y_i - f(x_i)
\big)^2 + \Pen(f) \Big \}.
\end{align}
In general, procedures of this type are referred to as non-parametric
$M$-estimators; in this paper, we are analyzing the case of
non-parametric and penalized least-squares.

Let us summarize some shorthand notation that we use throughout the
remainder of this paper.  For a pair of real-valued functions $f$ and
$g$ defined on the covariate space $\Xspace$, we define the squared
norm
\begin{subequations}
\begin{align}
\empnormsq{f - g} & \defn \frac{1}{\numobs} \sum_{i=1}^\numobs \big(
f(x_i) - g(x_i) \big)^2
\end{align}
along with the associated inner product
\begin{align}
\empinner{f}{g} & \defn \FULLSUM f(x_i) g(x_i).
\end{align}
\end{subequations}


\subsection{Excess risk and optimism}
\label{SecExcess}
Given a predictor $\fhat$, recall the
definition~\eqref{EqnDefnPopExcRisk} of its population excess risk.
Given the fixed design set-up of this paper, this population excess
risk takes the form
\begin{subequations}
\begin{align}
\label{EqnGoodExcess}  
  \ExPop(\fhat) & = \underbrace{\frac{1}{\numobs} \sum_{i=1}^\numobs
    \big(\fhat(x_i) - \fstar(x_i) \big)^2}_{ \equiv \|\fhat -
    \fstar\|_\numobs^2},
\end{align}
corresponding to the \emph{mean-squared error} or MSE between $\fhat$
and $\fstar$ evaluated on the fixed covariates
$\{x_i\}_{i=1}^\numobs$.

In analogy to the population excess risk~\eqref{EqnDefnPopExcRisk},
given a data set $\{(x_i, y_i) \}_{i=1}^\numobs$, we can also define
the \emph{empirical excess risk} as
\begin{align}
\ExHat(\fhat) & \defn \frac{1}{\numobs} \sum_{i=1}^\numobs (\resp_i -
\fhat(\covariate_i))^2 - \frac{1}{\numobs} \sum_{i=1}^\numobs
(\resp_i- \fstar(\covariate_i))^2.
\end{align}
\end{subequations}
In contrast to the population case, the empirical excess risk need not
be non-negative.

The difference between the population and empirical excess risks
defines the \emph{optimism} of an estimate $\fhat$.  For the
least-squares objective considered here, some simple algebra yields
\begin{align}
\label{EqnDefnOptimism}  
\ExPop(\fhat) & = \ExHat(\fhat) + 2 \, \Opt(\fhat) \quad \mbox{where}
\quad \Opt(\fhat) \defn \underbrace{\frac{1}{\numobs}
  \sum_{i=1}^\numobs \wnoise_i \big(\fhat(x_i) - \fstar(x_i) \big)}_{
  \equiv \empinner{\wnoise}{\fhat - \fstar} },
\end{align}
where $\wnoise_i \defn \resp_i - \fstar(\covariate_i)$ is the
effective noise in the regression model.  Thus, in order to obtain an
upper bound on the population excess risk, it suffices to obtain an
upper bound on the optimism $\Opt(\fhat)$.


\subsection{Wild refitting}
\label{SecWildRefit}
The main contribution of this paper is to demonstrate the
effectiveness---both empirically and theoretically---of a simple
refitting procedure for computing upper bounds on the optimism, and
hence the population excess risk.  As noted earlier, we refer to this
method as \emph{wild refitting}, since like one variant of the wild
bootstrap~(e.g.,~\cite{wu1986jackknife,liu1988bootstrap,li1989resampling,
  mammen1992bootstrap}), it makes use of random signs (Rademacher
variables) to symmetrize an appropriate set of residuals, and also
allows for general heteroskedasticity.  Note that $\Opt(\fhat)$ is a
random variable, since it depends on both the noise terms $\wnoise_i$
and estimate $\fhat$ itself.  Thus, the wild refitting procedure also
returns a random variable.

While our theoretical analysis focuses on $M$-estimators of the
form~\eqref{EqnMest}, the procedure itself applies to any statistical
method $\Meth$ that uses an $\numobs$-vector $y$ of responses to
compute a predictor $\fhat \defn \Meth(y)$ belonging to some function
space $\Fclass$.  Any instantiation of the wild refitting procedure
depends on a recentering predictor $\ftil: \Xspace \rightarrow \real$
and a scalar noise level $\nstd > 0$.  It requires only black-box
access to the estimator $\Meth$, that we can compute the function
$\Meth(u) \in \Fclass$ for any possible response vector $u \in
\real^\numobs$. \\

\begin{tcolorbox}[colback=gray!20, colframe=gray!60, arc=5mm, boxrule=1pt, width=\linewidth,
    title={\textcolor{black}{Wild Refitting}}, fonttitle=\bfseries]
  \begin{tabular}{ll}
    {\bf{Inputs:}} & Prediction method $\Meth: \real^\numobs
    \rightarrow \Fclass$ \\ & Response vector $y \in \real^\numobs$
    and fitted predictor $\fhat = \Meth(y)$ \\ & Recentering predictor
    $\ftil \in \Fclass$ and noise scale $\nstd > 0$.
  \end{tabular}
\begin{subequations}
\begin{enumerate}
\item[(i)] With the given recentering function $\ftil$, compute the residuals
  \begin{align}
\wtil_i & \defn y_i - \ftil(x_i) \qquad \mbox{for $i = 1, 2, \ldots,
  \numobs$.}
  \end{align}
\item[(ii)] Using original predictor $\fhat$ and given noise scale
  $\nstd > 0$, form the wild responses
  \begin{align}
   \ywild_i & \defn \fhat(x_i) + \underbrace{\nstd \rade{i}
     \wtil_i}_{\revdefn \wwild_i} \qquad \mbox{for $i = 1, \ldots,
     \numobs$},
  \end{align}
  where $\{ \rade{i} \}_{i=1}^\numobs$ is an i.i.d. sequence of
  Rademacher variables.
\item[(iii)] Compute the refitted wild solution $\fwild \defn \Meth(\ywild)$. 
\end{enumerate}
\end{subequations}  
\noindent \textbf{Outputs:}
\[
\left.
\begin{minipage}{0.6\linewidth}
\begin{itemize}
    \item Wild refit $\fwild$
    \item Wild noise vector $\wwildrho = (\wwild_i)_{i=1}^\numobs$
  \end{itemize}
\end{minipage}
\right\} 
\quad  \Longrightarrow 
\parbox[c]{0.3\linewidth}{Use to bound  excess risk \\
  $\ExPop(\fhat)$ via suitable choice of $\rho$}
\]
\end{tcolorbox}
\noindent In the simplest instantiation, we use the estimate $\fhat$
itself as the centering function (i.e., $\ftil \equiv \fhat$), but as
our theory clarifies, other choices of the centering function $\ftil$
can yield superior performance.  Moreover, the noise scale $\nstd$
plays an important role, and our theory provides guidance on its
choice.

The wild-refitting problem can be used to define an optimism term,
given by
\begin{align}
\label{EqnOptWild}  
\OptWild(\fwild) \defn
\frac{1}{\numobs} \sum_{i=1}^\numobs \rade{i} \wtil_i \big(
\fwild(x_i) - \fhat(x_i) \big).
\end{align}
In contrast to the true optimism $\Opt(\fhat)$, this wild optimism
depends only quantities that can be computed based on the original
observation vector $y \in \real^\numobs$, along with the Rademacher
sequence $\rade{} \in \{-1, +1 \}^\numobs$ introduced as part of the
wild-refitting procedure.  The main results of this paper are to
establish conditions under which the quantities defining this wild
optimism can be used to bound the true optimism, and hence the true
excess risk.

Notably---and in sharp contrast to asymptotic or distributional
guarantees provided by the wild bootstrap---we will provide guarantees
that hold based on a \emph{single realization} of the Rademacher
vector $\rade{} \in \{-1, +1\}^\numobs$.  Thus, we do not provide
distributional guarantees, but instead a non-asymptotic guarantee.
Moreover, the wild refitting procedure only requires a single call to
the prediction method $\Meth$, as laid out in item (d) of our
motivating desiderata (see~\Cref{SecContributions}).


\subsection{A simple illustration:  Tikhonov versus TV regularization}
\label{SecTik}

So as to provide intuition, let us begin by considering a simple form
of model selection in the context of non-parametric regression.  Let
$\fstar:[0, 1] \rightarrow \real$ be the unknown regression function,
and suppose that we collect $\numobs$ noisy observations of the form
\begin{align}
\label{EqnEquispace}  
  y_i = \fstar(x_i) + w_i \qquad \mbox{where $x_i = i/\numobs$ for $i
    = 1, \ldots, \numobs$,}
\end{align}
and $\{w_i\}_{i=1}^\numobs$ is an i.i.d. sequence of zero-mean noise
variables.  Various classical approaches are based on $M$-estimators
of the form~\eqref{EqnMest} with suitable penalty functions. Here we
consider two methods based on penalties from the family
\begin{align*}
  \Pen_q(f) & = \sum_{i=1}^{\numobs-1} \big| f(x_{i+1}) - f(x_i)
  \big|^q \qquad \mbox{for an exponent $q \geq 0$.}
\end{align*}
The choice $q = 2$ corresponds to \emph{Tikhonov regularization},
whereas the choice $q = 1$ corresponds to a \emph{total variation}
(TV) method.

\begin{figure}[ht!]
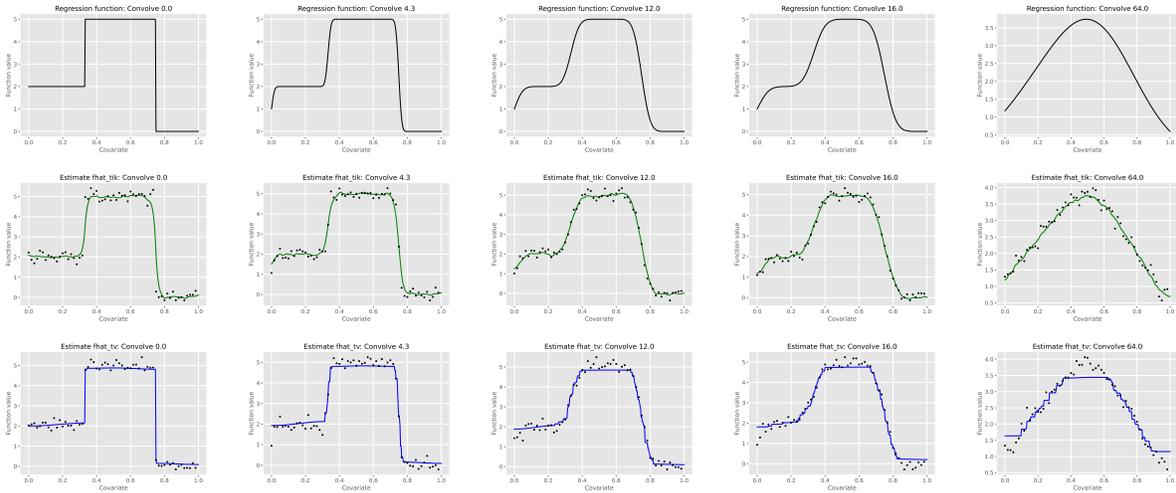

  \begin{center}
  \begin{tabular}{ccccc}
    \widgraph{0.17\textwidth}{\figdir/fig_gsmooth_0} &
    \widgraph{0.17\textwidth}{\figdir/fig_gsmooth_430} &
    \widgraph{0.17\textwidth}{\figdir/fig_gsmooth_1195} &
    \widgraph{0.17\textwidth}{\figdir/fig_gsmooth_1600} &
    \widgraph{0.17\textwidth}{\figdir/fig_gsmooth_6400} \\
    \widgraph{0.17\textwidth}{\figdir/fig_fhat_tik_0} &
        \widgraph{0.17\textwidth}{\figdir/fig_fhat_tik_430} &
    \widgraph{0.17\textwidth}{\figdir/fig_fhat_tik_1195} &
    \widgraph{0.17\textwidth}{\figdir/fig_fhat_tik_1600} &
        \widgraph{0.17\textwidth}{\figdir/fig_fhat_tik_6400} \\
        \widgraph{0.17\textwidth}{\figdir/fig_fhat_tv_0} &
    \widgraph{0.17\textwidth}{\figdir/fig_fhat_tv_430} &
    \widgraph{0.17\textwidth}{\figdir/fig_fhat_tv_1195} &
    \widgraph{0.17\textwidth}{\figdir/fig_fhat_tv_1600} &
    \widgraph{0.17\textwidth}{\figdir/fig_fhat_tv_6400} 
    \\
  \end{tabular}
  \end{center}
    \caption{Top row: illustration of the family of functions
      $\fstar_\gamma$ for different choices of the smoothing parameter
      $\gamma \geq 0$.  Middle row: Corresponding function estimate
      $\fhat$ with Tikhonov regularization.  Bottom row: corresponding
      function estimates with TV regularization.  }
  \label{FigSignals}
\end{figure}

The TV estimator imposes a sparsity constraint on the function
differences, and so is well-suited to reconstructing functions with
step-like behavior.  On the other hand, Tikhonov regularization
imposes a type of smoothness penalty, and so is better suited to
smoother functions.  So as to illustrate these trade-offs, we
constructed a family of functions that vary from step-like to smooth.
We began by constructing a function $\fstar_0: [0,1] \rightarrow
\real$ that is piecewise constant, with three distinct pieces, as
illustrated in the left-most panel in the top row
of~\Cref{FigSignals}.  By convolving this function with a zero-mean
Gaussian with variance $\gamma^2$, we obtain a sequence of
progressively smoother functions $\fstar_\gamma$, as shown in the top
row panels moving left to right.  Thus, the left-most function is best
suited to TV regularization, whereas the right-most function is best
suited to Tikhonov regularization.

\begin{figure}[h!]
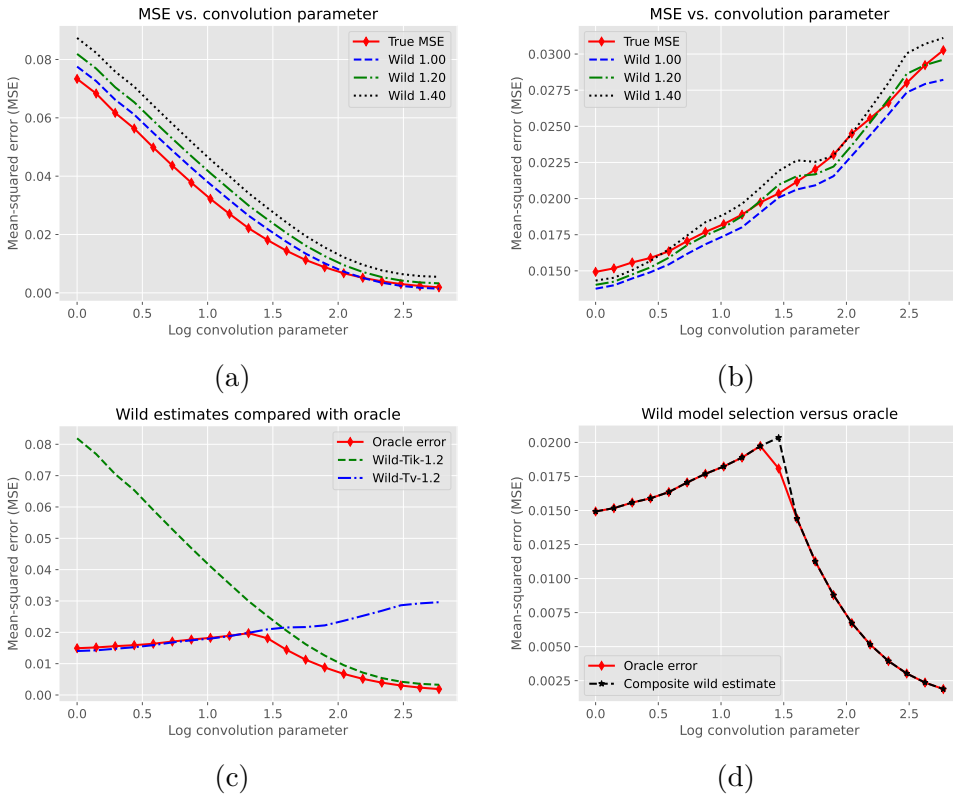

  \begin{center}
  \begin{tabular}{cc}
    \widgraph{0.38\textwidth}{\figdir/fig_full_tvsim_nx_301_sig_30_fhat_tik} &
        \widgraph{0.38\textwidth}{\figdir/fig_full_tvsim_nx_301_sig_30_fhat_tv} \\
        (a) & (b) \\
    \widgraph{0.38\textwidth}{\figdir/fig_full_tvsim_nx_301_sig_30_opt} &
   \widgraph{0.38\textwidth}{\figdir/fig_full_tvsim_nx_301_sig_30_wild_best} \\
        (c) & (d)
  \end{tabular}
  \end{center}
\caption{
    Panels (a) and (b):
    Comparison of the true MSE $\ExPop(\fhat)$ and wild estimates at
    noise scales $\rho \in \{1.0, 1.2, 1.4 \}$ for the Tikhonov
    estimate (panel (a)) and the TV estimate (panel (b)).  (c) Plots
    of the MSE obtained by an oracle (red diamonds) that can choose
    optimally between the two estimators.  (d) Comparison of model
    selection using the wild estimates to the oracle error.}
  \label{FigTikTV}
\end{figure}

We then applied both the Tikhonov and the TV estimates to observations
from the model~\eqref{EqnEquispace} for the $\gamma$-convolved
function $\fstar_\gamma$ over a range of convolution strengths
$\gamma$.  Panel (a) of~\Cref{FigTikTV} plots the excess risk
$\ExPop(\fhat)$ of the Tikhonov estimate versus $\gamma$, and compares
it to the wild estimates of excess risk for wild noise scales $\rho
\in \{1.0, 1.2, 1.4 \}$.  Panel (b) provides these same plots for the
TV estimate.  Consistent with the smoothness/sparsity interpretation,
note that the MSE of the Tikhonov method decreases as $\gamma$
increases, whereas that of the TV method exhibits the opposite
behavior.  Note also how the wild-estimated MSEs track this behavior.

In panel (c) of~\Cref{FigTikTV}, we plot (in red diamonds) the MSE
obtained by the \emph{oracle selector} that chooses, for each value of
$\gamma$, between the TV and Tikhonov estimate so as to minimize the
MSE. We refer to the MSE obtained in this as the oracle MSE. We also
overlay the wild estimates of the MSEs of these two methods.  In panel
(d), we plot the oracle MSE (red diamonds) versus the MSE obtained by
wild model selection (black stars).  The latter method performs
model selection based on the wild-estimated MSEs shown in panel (c).


\section{Analysis and non-asymptotic guarantees}
\label{SecMain}
We now turn to some analysis of the wild refitting procedure, in
particular in the form of non-asymptotic guarantees on the wild
optimism.  In this section, we prove two main results:
\begin{itemize}
\item \Cref{ThmWildOptBound} gives an upper bound on the true optimism
  in terms of the wild optimism $\OptWild(\fwild)$ evaluated at an
  appropriate wild noise scale $\rho$.  However, this choice depends
  on the estimation error $\rhat \defn \|\fhat - \fproj\|_\numobs$,
  which is unknown.
\item \Cref{ThmRhatError} closes this gap by showing how wild
  refitting can be used to generate a high-probability upper bound on
  $\rhat$.  This bound provides guidance on the setting of wild
  noise scale for which the guarantee from~\Cref{ThmWildOptBound}
  holds.
\end{itemize}

\subsection{Problem set-up and wild complexity}

Our analysis applies to an estimate $\fhat$ generated by solving an
optimization problem of the form~\eqref{EqnMest}. Given this
structure, it is natural to define
  \begin{align}
\label{EqnDefnFproj}    
\fproj & = \arg \min_{f \in \Fclass} \Big \{ \empnorm{f - \fstar}^2 +
\Pen(f) \Big \},
\end{align}
corresponding to the best penalized approximation to $\fstar$ in the
absence of any noise.  Note that $\fproj$ is a deterministic function,
since we are viewing the covariates as fixed.

\subsubsection{Assumptions}

We now describe the assumptions that underlie our analysis.

\paragraph{Firm non-expansiveness:}
For a given function $f \in \Fclass$, introduce the convenient
shorthand \mbox{$f(x_1^n) \defn \big(f(x_1), \ldots, f(x_n) \big) \in
  \real^\numobs$.}  In the bulk of our analysis, we focus on penalty
functions $\Pen: \Fclass \rightarrow \real^+$ of the form
\begin{align}
\Pen(f) & \defn \begin{cases} 0 & \mbox{if $f(x_1^n) \in \Convex$,
    and} \\
  + \infty & \mbox{otherwise,}
\end{cases}
\end{align}
where $\Convex \subset \real^\numobs$ is a given compact subset of
$\real^\numobs$.  In future work, we will explore theoretical
extensions to regularized estimators in the Lagrangian case.

Our analysis requires that the resulting method $\Meth: \real^\numobs
\rightarrow \Fclass$, as defined in by equation~\eqref{EqnMest}, be
\emph{firmly non-expansive} around $\fstar$.  More precisely, noting
that $\fproj = \Meth(\fstar(x_1^n))$ by assumption, we require that
the estimate $\ftil = \Meth(\fstar(x_1^n) + u)$ obtained from via any
perturbation vector $u \in \real^\numobs$ satisfies the inequality
\begin{subequations}
\begin{align}
  \label{EqnFirm}
\underbrace{\|\Meth \big(\fstar(x_1^n) + u \big) -
  \Meth\big(\fstar(x_1^n) \big) \|_\numobs^2}_{\equiv \|\ftil -
  \fproj\|_\numobs^2} & \; \; \leq \; \; \inprod{u}{\ftil -
  \fproj}_\numobs.
\end{align}
When the set $\Convex$ is convex, then this firm-expansiveness follows
from standard optimization-theoretic properties of projections onto
convex sets (e.g.,~\cite{HirLem01}).  It is worth noting, however,
that a local property of this type can hold more generally for certain
non-convex sets (e.g., prox-regular sets~\cite{RocWet09}).

\paragraph{Noise conditions:}  Recall that our observations
take the form $y_i = \fstar(x_i) + w_i$, where each $w_i$ is an
additive noise variable.  We
require that
\begin{align}
\label{EqnNoiseCond}  
\mbox{\underline{Conditioned on $\{x_i\}_{i=1}^\numobs$:}} \qquad
\mbox{$(w_1, \ldots, w_n)$ are independent, each with symmetric
  distribution.}
\end{align}
\end{subequations}
The conditional symmetry assumption implies that $\Exs[w_i \mid x_i] =
0$, a fact that we exploit in our analysis.  When the symmetry
condition fails to hold, the bounds that we give here all hold with an
increased constant factor.  (Essentially, we lose a factor of two in
order to handle non-symmetric noise, since we need to introduce two
separate copies of $w$ and $w'$, and take their difference in order to
form the symmetrized analog.) \\

\noindent {\bf{Remark:}} Other than the
conditions~\eqref{EqnNoiseCond}, we impose no other assumptions on the
noise variables.  In particular, they are permitted to be
heteroskedastic and/or exhibit highly non-Gaussian tail behavior.

\subsubsection{From wild optimism to wild complexity}

For future reference, it is convenient to highlight an important
connection between the wild optimism, and a function $r \mapsto
\Zwild(r)$ that we refer to as the wild complexity.  Recall that the
\emph{wild optimism} at noise scale $\rho$ for the estimate $\fhat$ is
given by
\begin{subequations}
\begin{align}
  \label{EqnOptWildTwo}
\OptWild(\fwild) \defn \frac{1}{\numobs} \sum_{i=1}^\numobs \rade{i}
\wtil_i \big( \fwild(x_i) - \fhat(x_i) \big), \qquad \mbox{where
  $\wtil_i = y_i - \ftil(x_i)$.}
\end{align}
Here $\ftil$ is the pilot estimator (frequently equal to $\fhat$), and
$\fwild = \Meth(\ywild)$ is the method applied to the wild
observations $\ywild_i = \fhat(x_i) + \rho \rade{i} \wtil_i$.  On the
other hand, the \emph{wild noise complexity} at $\fhat$ is the
function given by
\begin{align}
\label{EqnDefnZwild}  
r \mapsto \Zwild(r) & \defn \sup_{f \in \MyBall{\fhat}{r}} \Big[
  \FULLSUM \rade{i} \noisetil_i \big(f(x_i) - \fhat(x_i) \big) \Big].
\end{align}
\end{subequations}
This wild noise complexity is intimately related to the wild optimism,
as formalized by the following result:
\begin{lemma}[From wild optimism to wild complexity]
\label{LemOpt2Complex}  
For any wild noise scale $\rho > 0$, we have
\begin{align}
\label{EqnOpt2Complex}  
  \OptWild(\fwild) & = \Zwild(\|\fwild - \fhat\|_\numobs).
\end{align}
\end{lemma}
\noindent See~\Cref{SecLemOpt2Complex} for the proof of this claim. \\

\subsection{Bounding the true optimism}

With these ingredients in place, we are now ready to state a
high-probability bound on the true optimism $\Opt(\fhat)$ in terms of
the wild optimism $\OptWild(\fwild)$.

\grayboxed{
\begin{theorem}[Wild optimism bound]
\label{ThmWildOptBound}  
Under the noise condition~\eqref{EqnNoiseCond}, consider any radius
$r$ such that \mbox{$\empnorm{\fhat - \fproj} \leq r$,} and let $\rho
> 0$ be the noise scale for which $\|\fwild - \fhat\|_\numobs = 2 r$.
Then for any $\tpar > 0$, we have
\begin{align}
\label{EqnWildOptBound}
\Opt(\fhat) \leq \OptWild(\fwild) \; + \; \big \{
\HighPlain_\numobs(\tpar) + \AppError \big \} \quad \mbox{with
  prob. at least $1 - 4 e^{-t^2}$,}
\end{align}
where the probability deviation term is given by
\begin{subequations}
  \begin{align}
\label{EqnDefnHigh}    
\HighPlain_\numobs(\tpar) \defn \big \{ 3 r + \empnorm{\fproj -
  \fstar} \big \} \DevTerm
\end{align}
and the pilot error is given by
\begin{align}
\label{EqnDefnPilot}  
\AppError \defn \sup \limits_{f \in \MyBall{\fhat}{2r}} \FULLSUM
\rade{i} (\ftil(x_i) - \fstar(x_i)) \: (f(x_i) - \fhat(x_i)).
\end{align}
\end{subequations}
\end{theorem}
}
\noindent See~\Cref{SecThmWildOptBound} for the proof of
this theorem. \\

To interpret this result, note that inequality~\eqref{EqnWildOptBound}
shows that the true optimism $\Opt(\fhat)$ can be upper bounded by the
wild optimism for an \emph{appropriately chosen} $\rho$, along with
the probability deviation term~\eqref{EqnDefnHigh} and the pilot error
term~\eqref{EqnDefnPilot}.  Let us comment on these two terms.

\paragraph{Probability deviation term:}
The guarantee holds with exponentially high probability (i.e.,
\mbox{$1 - 4 e^{-\tpar^2}$}) in the deviation term $\tpar$; this sharp
tail behavior arises due to the highly favorable concentration
behavior of Rademacher variables used the wild refitting procedure.
Our proof exploits concentration inequalities for convex and Lipschitz
functions of Rademacher variables so as to preserve this behavior. The
function $\HighPlain_\numobs(\tpar)$ defined in
equation~\eqref{EqnDefnHigh} scales as $t
\|w\|_\infty/\sqrt{\numobs}$, so that the distributional properties of
the additive noise vector $w \in \real^\numobs$ enter purely via the
sup-norm $\|w\|_\infty$.  This fact means that it is possible to
handle noise with tail behavior much heavier than Gaussian.  For
example, consider a noise distribution that has only a finite
polynomial moment---say $\Exs|w|^k \leq C$ for some $k > 2$.  An easy
calculation shows that $\Exs\|w\|_\infty \leq (C \numobs)^{1/k}$, with
similar bounds for the event $\{\|w\|_\infty \geq s \}$, so
that~\Cref{ThmWildOptBound} can still yield non-trivial guarantees.
We illustrate the behavior of the wild-refitting procedure for
heavy-tailed Student $t$-noise in~\Cref{SecPlug}; in particular,
see~\Cref{FigStudent}.

\paragraph{Pilot error term:}  Now let us consider the pilot error
term~\eqref{EqnDefnPilot}, and in particular its relation to the wild
optimism.  From the equivalence~\eqref{EqnOpt2Complex}, the wild
optimism in the bound~\eqref{EqnWildOptBound} can be written as
\begin{align*}
\OptWild(\fwild) & = \Zwild(2 r) \; = \; \sup \limits_{f \in
  \MyBall{\fhat}{2r}} \FULLSUM \rade{i} \wtil_i (f(x_i) - \fhat(x_i)).
\end{align*}
By comparison with the definition~\eqref{EqnDefnPilot}, we see that
the pilot error has the same form, with the residual $\wtil_i$
replaced by the difference $d(x_i) \defn \ftil(x_i) - \fstar(x_i)$.
Consequently, as long as the pilot function $\ftil$ is a reasonable
estimate $\fstar$---so that the difference $d(x_i)$ is typically
smaller than the noise $w_i$---then we can expect that the pilot error
is dominated by $\OptWild(\fwild)$.


\subsection{Bounding the estimation error}

The key assumption in \Cref{ThmWildOptBound} is the existence of some
known $r$ such that $r \geq \|\fhat - \fstar\|_\numobs$.  Thus, it
becomes useful only in so far as we can obtain an upper bound on the
\emph{statistical estimation error} $\rhat \defn \empnorm{\fhat -
  \fproj}$.  This quantity corresponds to the error associated with
estimating the best approximation $\fproj$ to $\fstar$ from the chosen
function class $\Fclass$ (cf. equation~\eqref{EqnDefnFproj}).  In this
section, we describe an explicit procedure that can be used to obtain
upper bounds on $\rhat$.  It involves an inequality defined in the
terms of the wild complexity $\Zwild(r)$ from
equation~\eqref{EqnDefnZwild}.

\grayboxed{  
\begin{theorem}[Estimation error bounds via wild complexity]
\label{ThmRhatError}
Under the non-expansive condition~\eqref{EqnFirm} and the noise
condition~\eqref{EqnNoiseCond}, for any $t \geq 3$, we have
\begin{subequations}  
\begin{align}
\label{EqnRhatError}  
\rhat^2 & \leq \max \Big \{\Zwild \big( \big(\twoplust \big)
\rhat\big), \; \: \tfrac{\tpar^4}{\numobs} \Big \} + \HackHigh +
\AppError
\end{align}
with probability at least $1 - 2 e^{-t^2}$.  Moreover, if $\Fclass$ is
convex, then with the same probability, for any noise scale $\rho >
0$, we have
\begin{align}
\label{EqnWildRefit}  
  \rhat^2 & \leq \max \Big \{ (\rwild)^2, \; \: \tfrac{\rhat}{\rwild}
  \: \Zwild \big( \,(\twoplust) \, \rwild \big), \;
  \tfrac{\tpar^4}{\numobs} \Big \} + \HackHigh + \AppError,
\end{align}
\end{subequations}
where $\rwild = \|\fwild - \fhat\|_\numobs$ is the error of
the wild refit $\fwild$.
\end{theorem}
}
\noindent See~\Cref{SecThmRhatError} for the proof of this claim. \\

Let us make a few comments to interpret the meaning of the two bounds
in the theorem.  Both bounds involve the pilot error term $\AppError$,
as previously discussed, along with probability deviation parameter
$\tpar$ and sup-norm $\|w\|_\infty$ of the noise vector.  Focusing on
the bound~\eqref{EqnRhatError}, the $\tpar^4/\numobs$ term within the
maximum on the right-hand side is of lower order whenever $\rhat^2
\succsim 1/\numobs$, which is the case in all but the simplest
parametric problem.  Disregarding these terms, the rough
interpretation is that we have an upper bound of the form $\rhat \leq
r$ for any radius $r$ such that
\begin{align}
  r^2 \geq \Zwild \big( \big(\twoplust \big) r \big).
\end{align}
Finding the best such bound amounts to finding the smallest radius $r$
that satisfies this inequality.  Note that this can be done, because
for a given radius $s$, the wild complexity $\Zwild(s)$ is computable
based purely on the given estimate $\fhat$, along with the wild noise
variables $\{\wtil_i \}_{i=1}^\numobs$.  Alternatively, from the
equivalence~\eqref{EqnOpt2Complex} from~\Cref{LemOpt2Complex}, we can
compute $\Zwild(s)$ by varying the wild noise scale $\rho$ until the
wild estimate $\fwild$ satisfies $\|\fwild - \fhat\|_\numobs = s$.  On
the other hand, the bound~\eqref{EqnWildRefit} is a slightly weaker
but more interpretable result that holds for convex function classes.
We discuss its use and geometric meaning below.


\subsection{Putting together the pieces}

Let us now summarize the overall conclusions that can be drawn
from~\Cref{ThmWildOptBound,ThmRhatError} in conjunction, focusing on
the case of a convex function class $\Fclass$.  Consider the problem
of upper bounding $\rhat = \|\fhat - \fproj\|_\numobs$.  If we
disregard the deviation and approximation terms, then a rough summary
of the bound~\eqref{EqnWildRefit} is via the inequality
\begin{align}
\label{EqnMasterBound}  
\rhat & \leq \max \Big \{ \rwild, \tfrac{\Zwild \big(2 \rwild
  \big)}{\rwild} \Big \} \; \stackrel{(\dagger)}{\leq} \;
\underbrace{\max \Big \{ \rwild, \tfrac{2 \Zwild \big(\rwild
    \big)}{\rwild} \Big \}}_{\equiv B(\rho)}
\end{align}
Here inequality $(\dagger)$ follows from the concavity of the function
$u \mapsto \Zwild(u)$, which implies that $u \mapsto \Zwild(u)/u$ is
decreasing.  See~\Cref{LemZwildConcave} in~\Cref{SecCorWildRefit} for
details.

\begin{figure}[h!]
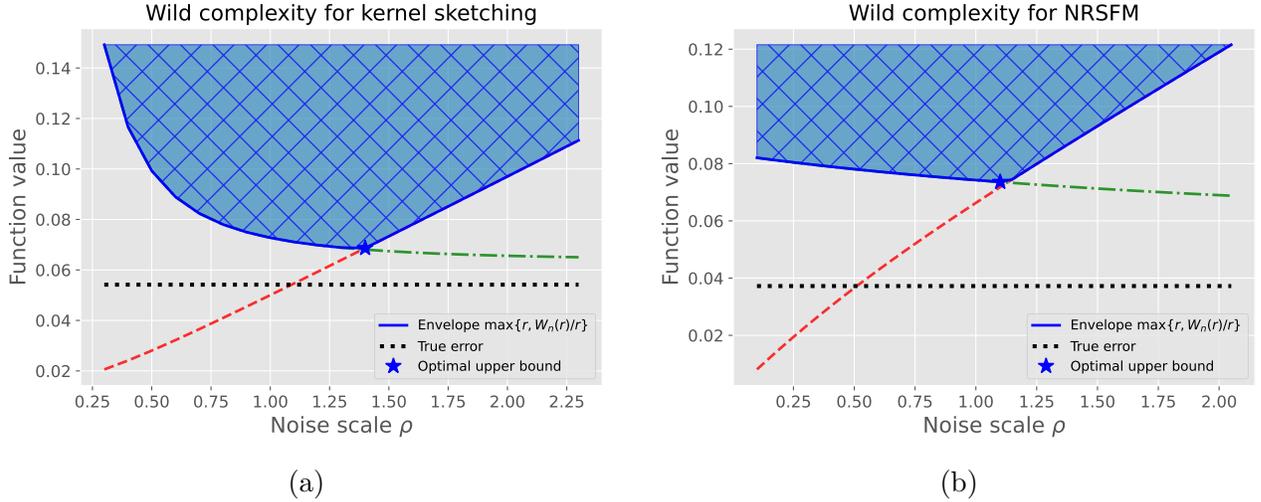

  \begin{center}
    \begin{tabular}{cc}
      \widgraph{0.5\textwidth}{\figdir/fig_sketch_wfun_sdim_550}&
            \widgraph{0.5\textwidth}{\figdir/fig_dance_wfun_reg_205} \\
      (a) & (b) 
    \end{tabular}
  \end{center}
  \caption{The family of inequalities defined by the
    functions~\eqref{EqnTwoFunctions} can be used to compute upper
    bounds on the true error $\rhat = \empnorm{\fhat - \fproj}$.
    Illlustrated for two different prediction problems: (a) Randomized
    sketching in kernel regression~\cite{YanPilWai17}.  (b) Non-rigid
    shape-from-motion (\Cref{SecDance}).}
  \label{FigMasterBound}
\end{figure}

Observe that the bound~\eqref{EqnMasterBound} holds for each choice of
noise scale $\rho$; consequently, we can optimize this choice so as to
obtain the tightest possible bound.  The function $\rho \mapsto
B(\rho)$ defined in equation~\eqref{EqnMasterBound} is the maximum of
the two functions
\begin{align}
  \label{EqnTwoFunctions}
  B_1(\rho) \defn \rwild, \quad \mbox{and} \quad B_2(\rho) \defn
  \frac{2 \Zwild(\rwild)}{\rwild} \qquad \mbox{where $\rwild \defn
    \|\fwild - \fhat\|_\numobs$.}
\end{align}
In~\Cref{FigMasterBound}, we plot these two functions versus $\rho$
for two different problems: (a) randomized sketching approximations
for kernel ridge regression~\cite{YanPilWai17}; and (b) non-rigid
structure-from-motion (see~\Cref{SecDance} for details on the latter
application).

The error $\rwild$ is an increasing function\footnote{Our use of
increasing does not mean strictly so; we are using increasing in the
sense of non-decreasing.}  of the noise scale $\rho$, so that the
function $B_1$ is increasing in $\rho$.  In~\Cref{FigMasterBound},
this increasing function is plotted with a red-dotted line.  On the
other hand, as noted above, the function $u \mapsto \Zwild(u)/u$ is
decreasing $u$, which implies that the function $B_2$ is decreasing in
the wild noise scale $\rho$.  This decreasing function is shown with a
dashed-dotted green line in~\Cref{FigMasterBound}.  The maximum of the
two functions (i.e., $\rho \mapsto \max \{B_1(\rho), B_2(\rho) \}$),
plotted in a solid blue line, provides an upper bound on $\rhat =
\|\fhat - \fproj\|_\numobs$.  We can find the optimal bound by varying
the noise scale $\rho$ up to the point where $B_1(\rwild) =
B_2(\rwild)$, or equivalently $\rwild = 2
\frac{\Zwild(\rwild)}{\rwild}$.  This optimal upper bound is marked in
the plots with a blue star.

Once we have obtained an upper bound on $\rhat$, we can then return
to~\Cref{ThmWildOptBound}.  It suggests that---apart from the
higher-order and pilot terms---we can upper bound the true optimism
$\Opt(\fhat)$ via the wild optimism $\OptWild(\fwild)$ for the noise
scale that we have chosen.  Finally, we combine this upper bound on
the optimism with equation~\eqref{EqnDefnOptimism} so as to upper
bound the excess risk.


\section{Some numerical studies}
\label{SecNumerical}

In this section, we illustrate the performance of wild refitting on
two different classes of problems:
\begin{itemize}
\item \Cref{SecDance}: we predict the excess risk for low-rank matrix
  estimators used for recovery in non-rigid shape-from-motion.  We
  illustrate how the wild refitting method can be used to select an
  appropriate rank for a standard method based on low-rank matrix
  approximation~\cite{Dai2014_nrsfm}.
\item \Cref{SecPlug}: we illustrate its behavior for plug-and-play
  denoising methods based on deep neural network priors.
\end{itemize}


\subsection{Non-rigid structure-from-motion}
\label{SecDance}

We begin by illustrating the wild refitting procedure in application
to the problem of non-rigid structure-from-motion (NRSFM) recovery,
which is a standard problem in computer vision with a lengthy history
(e.g., ~\cite{Bregler2000, Akhter2011_nrsfm, Dai2014_nrsfm}).  In our
illustration, we focus on a NRSFM recovery method based on low-rank
matrix approximation, as described by Dai et al.~\cite{Dai2014_nrsfm}.

For a given instance of NRSFM recovery, the dataset consists of
$\numframes$ observations of a collection of $\numpoints$ points.
Each frame is taken by a camera at a particular time and position, and
each point represents a marker associated with some type of non-rigid
moving object.  In the CMU MoCap dataset~\cite{cmumocap} used for this
investigation, each point corresponds to a marker on the moving
object.  The points lie in $3d$ space, but this $3d$-location is
unknown to us; instead, each frame of the camera sequence gives a
$2d$-projection of the unknown $3d$-position for each point.  By
suitable algebraic re-arrangement, the full dataset can be described
by a matrix $\Ymat \in \real^{2m \times p}$, where each point $j = 1,
\ldots, p$ is associated with a pair of rows in $\Ymat$.

Our goal is to recover the unknown \emph{shape matrix} $\SmatStar \in
\real^{3m \times p}$ that gives the $3d$-positions of each of the
$\numpoint$-points at each of the $m$ frames.  This unknown shape
matrix is related to the observation matrix $\Ymat \in \real^{2m
  \times p}$ via the linear relationship
\begin{subequations}
\begin{align}
\label{EqnNRSFMObserve}  
  \Ymat & = \Rmat \SmatStar + \Wmat,
\end{align}
where $\Rmat \in \real^{2m \times 3m}$ is a known camera-matrix for
the $3d$ to $2d$ transformation in each frame, and $\Wmat \in
\real^{2m \times p}$ in an additive noise term.

\begin{figure}[ht!]
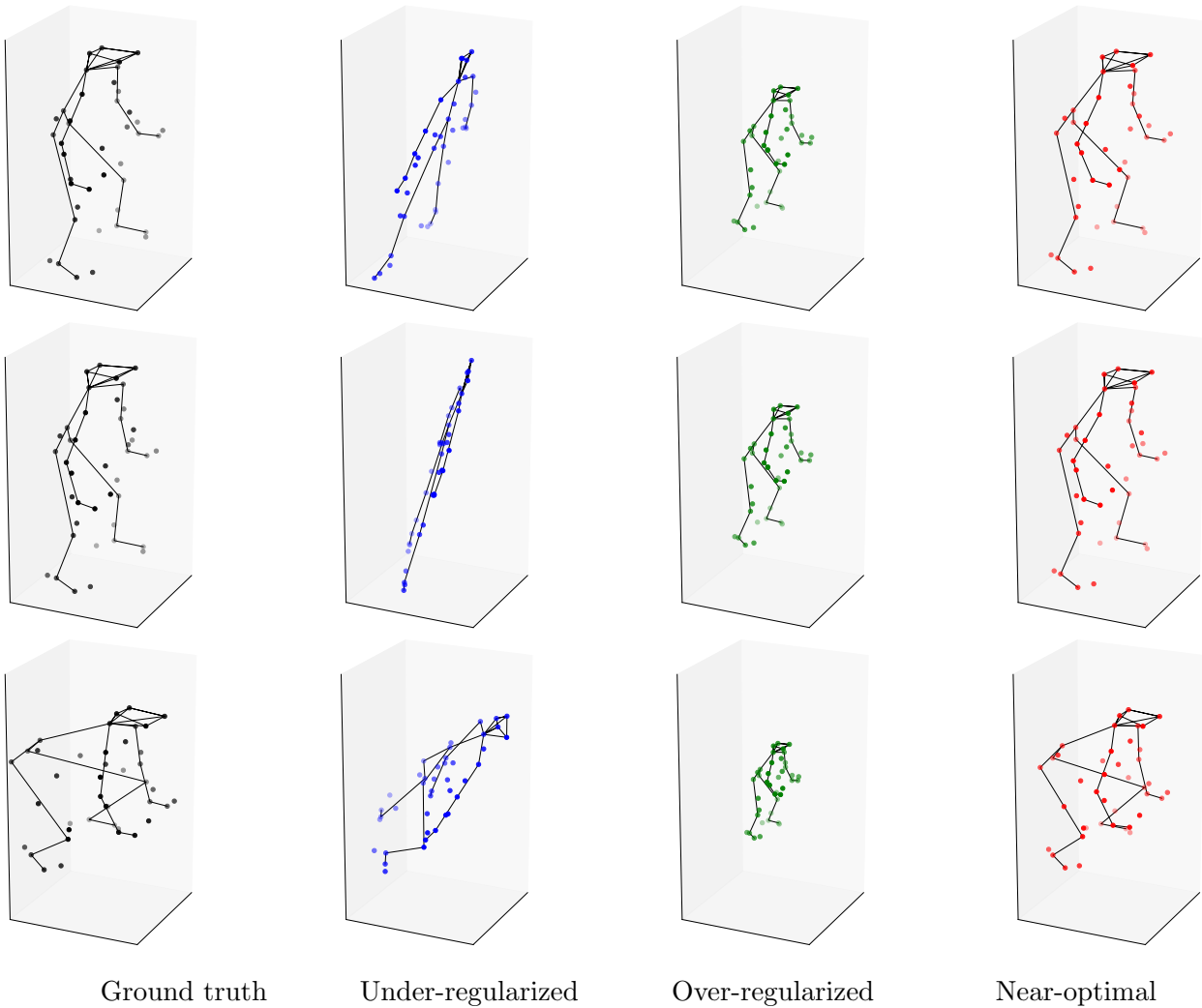

\begin{flushleft}
\begin{tabular}{c}
  \widgraph{1.0\textwidth}{\figdir/allfour-0}
  \\ \widgraph{1.0\textwidth}{\figdir/allfour-40}
  \\ \widgraph{1.\textwidth}{\figdir/allfour-80}
\end{tabular}
\begin{flushleft}
\qquad \qquad Ground truth \hspace*{0.4in} Under-regularized \quad
\qquad Over-regularized \qquad \qquad Near-optimal
\end{flushleft}
\end{flushleft}
\begin{center}
\caption{Comparison of reconstructed frames to ground truth for motion
  capture data (``Pick-up'' sequence from the CMU motion capture
  dataset~\cite{cmumocap}).  Reconstructions obtained by minimizing
  Frobenius norm subject to a nuclear norm
  constraint~\eqref{EqnLowRankEst} with three different choices of the
  radius $r$.  First column: Three frames of ground truth movement,
  showing the part of the ``Pick-up'' movement.  Second column:
  Under-regularized reconstruction (rank $r = 900$ is too
  large). Third column: Over-regularized reconstruction (rank $r =
  200$ is too small).  Fourth column: Reconstruction with rank $r =
  \exp(5.8) \approx 330$ chosen by the wild refitting procedure;
  see~\Cref{FigPickupMSE} and surrounding discussion for details.}
\label{FigPickup}
\end{center}
\end{figure}
Note that the observation matrix $\Ymat$ contains a total of $2 m p$
entries, whereas the unknown shape matrix $\SmatStar$ has $3 m p$
entries in total.  Consequently, the problem is ill-specified without
further restrictions, and it is essential to impose some form of
regularity condition on $\SmatStar$.  One widely-used approach is to
assume that the shape matrix $\SmatStar$ is relatively low-rank; this
modeling assumption can be interpreted as the shape matrix being
representable as a sum of a relatively small number of base shapes.

\begin{figure}[h]
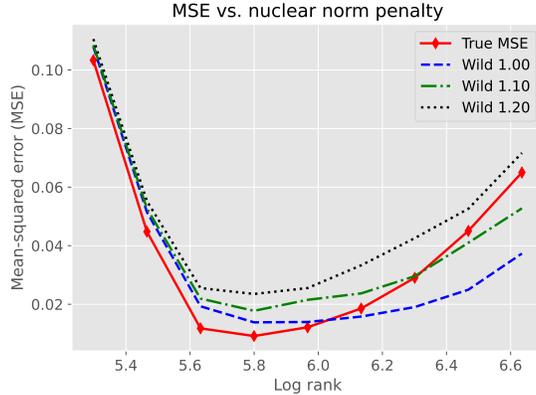

  \begin{center}
  \widgraph{0.45\textwidth}{\figdir/fig_cpickup_wild_mses_20}
  \end{center}
  \caption{Plots of the true MSE (red solid line with diamonds) versus
    the log radius ($\log r$), and comparison with wild refitting
    estimates of the MSE for three different choices of the noise
    scale $\rho \in \{1.0, 1.1, 1.2 \}$.  Consistent with our theory,
    a sufficiently large value of $\rho$ leads to an upper bound on
    the MSE.  We used the wild-estimated MSE with $\rho = 1.2$ to
    select the radius; doing so leads to the choice $r^* = \exp(5.8)
    \approx 330$.}
  \label{FigPickupMSE}
\end{figure}

Given such a low-rank assumption, a natural convex estimator---as
studied in the paper~\cite{Dai2014_nrsfm}---is to minimize a
data-fidelity term subject to a constraint on the nuclear norm
$\nucnorm{\Smat}$ of the shape matrix.  (The nuclear norm of a matrix
can be viewed as a convex surrogate to the rank function, which is
non-convex.)  This combination leads to the family of estimators
\begin{align}
  \label{EqnLowRankEst}
  \ShatEst_r \in \arg \min_{\Smat \in \real^{3 m \times p}}
  \frobnorm{\Ymat - \Rmat \Smat}^2 \qquad \mbox{such that
    $\nucnorm{\Smat} \leq r$},
\end{align}
\end{subequations}
where $\frobnorm{\cdot}$ denotes the Frobenius norm, and $r \geq 0$ is
the nuclear norm radius that we are free to choose. In practice, a
``good choice'' of $r$ is essential in obtaining accurate
reconstructions, and we made use of the wild refitting procedure in
order to make this choice, as we now describe.

\Cref{FigPickup} shows results obtained by applying the nuclear norm
constrained estimator~\eqref{EqnLowRankEst} to the ``Pick-Up''
sequence from the CMU MoCap database~\cite{cmumocap}.  It consists of
$p = 41$ points that are measured over a total of $m = 238$ frames.
The left column in~\Cref{FigPickup} plots three of these frames for
the ground truth dataset, which shows the basic form of the pick-up
movement.  (For clarity of the visualization, the plot also includes
additional lines marking the body, but these are not part of the
dataset itself.)

We then formed the data matrix $\Ymat$ as in
equation~\eqref{EqnNRSFMObserve}, where the noise matrix $\Wmat$ had
i.i.d. Gaussian entries with standard deviation $\sigma = 0.25$.  We
then solved the convex program~\eqref{EqnLowRankEst} for a wide range
of nuclear norm radii $r$.  The second, third and fourth panels
in~\Cref{FigPickup} plot three frames of these reconstructions.  The
second column uses radius $r = 900$, leading to an under-regularized
solution (stick figure spins around wildly), whereas the third column
uses radius $r = 200$, leading to an over-regularized solution (stick
figure simply shrinks). The final column shows three frames of the
reconstruction using the radius $r^* \approx 330$, which is the choice
suggested by our wild refitting procedure, which we describe next.

In order to choose the radius, we used the following version of wild
refitting. Fixing collection of nine radii $r \in [200, 900]$, we
performed the following steps for each $r$
\begin{itemize}
\item we first computed the estimate $\ShatEst_r$ in
  equation~\eqref{EqnLowRankEst};
\item then using this estimate, we computed the associated wild
  optimism $\OptWild(\fwild)$ for three different choices of
  wild noise scale---namely, $\rho \in \{1.0, 1.1, 2.2 \}$.
\end{itemize}
We then used these wild optimisms to construct surrogates to the true
optimism, and then formed approximations to the MSE via
equation~\eqref{EqnDefnOptimism}.

\Cref{FigPickupMSE} gives plots of the true MSE, along with these
three wild-refitting estimates of the MSE, versus the log radius $\log
r$ as a parameter.  We used the wild-refitted MSE with $\rho = 1.2$ to
choose the nuclear norm radius, and the minimum of this curve gives
the value $r^* = \exp(5.8) \approx 330$.  This choice $r^*$ was used
to produce the reconstruction $\ShatEst_{r^*}$ in the right-most
column of~\Cref{FigPickup}.


\subsection{Plug-and-play image restoration}
\label{SecPlug}

We now illustrate the use of wild refitting for MSE estimation for
image denoising.  The denoising problem is a core challenge in
statistical image processing, and has received renewed attention given
the central role of denoising in generative image modeling.

\begin{figure}[ht!]
  \begin{center}
    \begin{tabular}{cccccc}
      \widgraph{0.15\textwidth}{\figdir/fig_cropped_sloth_gauss_50_alg_125}
      &
      \widgraph{0.15\textwidth}{\figdir/fig_cropped_sloth_gauss_50_alg_100}
      &
      \widgraph{0.15\textwidth}{\figdir/fig_cropped_sloth_gauss_50_alg_80}
      &
      \widgraph{0.15\textwidth}{\figdir/fig_cropped_sloth_gauss_50_alg_30}
      &
      \widgraph{0.15\textwidth}{\figdir/fig_cropped_sloth_gauss_50_alg_10}
      &
      \widgraph{0.15\textwidth}{\figdir/fig_cropped_sloth_gauss_50_alg_0}
    \end{tabular}
        \begin{tabular}{c}
          \widgraph{0.5\textwidth}{\figdir/fig_mse_small_sloth_gauss_25} \\
          (a) \\
          \widgraph{0.8 \textwidth}{\figdir/fig_psnr_gauss_small_sloth_scale_130} \\
          (b) 
    \end{tabular}
  \end{center}
  \caption{ Performance of wild refitting for MSE estimation for
    plug-and-play denoising for Gaussian
    noise. \protect \input{denoise_caption.tex} Wild refitting with $\rho =
    1.15$ yields performance that is identical to the oracle in this
    particular case.}
  \label{FigGauss}
\end{figure}

\subsubsection{Set-up}

Let us begin by setting up the denoising problem using the notation of
this paper.  In abstract terms, we can represent an 2D-image as a
function $\fstar: [0,1]^2 \rightarrow \real$.  The digitized version
of the image---one that we would view on a computer---can be thought
of as a finite set of function evaluations: fixing some set of
covariates $\{x_i \}_{i=1}^\numobs$ contained within the unit square,
the digitized image is defined by the function evaluations
$\{\fstar(x_i) \}_{i=1}^\numobs$.  In a typical case, the covariates
would form a regular grid pattern on the unit square $[0,1]^2$.

In a denoising problem, we observe $\numobs$ samples of the form $y_i
= \fstar(x_i) + w_i$ for some additive noise variables $w_i$
satisfying the noise conditions~\eqref{EqnNoiseCond}, and our goal is
to recover an accurate estimate $\fhat$ of the original image
$\fstar$.  A broad class of methods are based on solving an
optimization problem of the form~\eqref{EqnMest}, for a suitably
penalty function $\Pen: \Fclass \rightarrow \real$.  Often, the
exponentiated quantity $e^{-\Pen(f)}$ is interpreted (modulo
normalization) as a prior over the space of images.  In these
approaches to image denoising---now known as plug-and-play
methods~\cite{venkatakrishnan2013plug,chan2016plug,graikos2022diffusion,hurault2022proximal}---the
function $\Pen$ (or its gradient $\nabla \Pen$) is learned from a
large image database, typically using deep neural networks.  This
penalty function is then combined with iterative optimization methods
(e.g., proximal gradient, ADMM etc.) so as to solve the underlying
optimization problem~\eqref{EqnMest}.  We refer the reader to Kamilov
et al.~\cite{kamilov2023plug} for a survey overview.

In order to explore the use of wild refitting for plug-and-play
denoising, we used a Python implementation of a proximal plug-and-play
method due to Hurault et al.~\cite{hurault2022proximal}. It provides a
sequence of penalty functions $\{ \Pen_\gamma, \gamma \geq 0 \}$,
indexed by the \emph{smoothness parameter} $\gamma$ (corresponding to
noise level used in training the deep net prior).  In our experiments,
we explore the use of wild refitting to automatically select this
tuning parameter so as to minimize the MSE of the resulting
reconstruction $\fhat \equiv \fhat_\gamma$.

\paragraph{Using wild refitting for smoothing parameter selection:}
In practice, given a noisy image $y$, we would like an automated way
of choosing the bandwidth parameter so that the estimate $\fhat_\gamma
= \Meth_\gamma(y)$ has smallest MSE.  The \emph{oracle choice} is to
choose
\begin{subequations}
\begin{align}
  \label{EqnOracleGamma}
  \gamma^* \defn \arg \min_{\gamma \geq 0} \|\fhat_\gamma -
  \fstar\|_\numobs^2.
\end{align}
However, it is not implementable, since the true image $\fstar$ is
unknown.  However, for a suitable choice of $\rho$, wild refitting
gives us an upper bound $B_\rho(\gamma)$ on the MSE $\|\fhat_\gamma -
\fstar\|_\numobs^2$, and so we can attempt to mimic the oracle rule by
instead choosing the value
\begin{align}
  \label{EqnWildChoice}
  \gamwild \defn \arg \min_{\gamma \geq 0}
  B_\rho(\gamma).
\end{align}
\end{subequations}
We can then compare the oracle MSE $\|\fhat_{\gamma^*} -
\fstar\|_\numobs^2$ to the MSE $\|\fhat_{\gamwild} -
\fstar\|_\numobs^2$ obtained from the wild refitting choice
$\gamwild$.

\subsubsection{Results}

We performed experiments for various images, and various types of
noise, including Gaussian, Student-$t$ noise, and heteroskedastic
ensembles.  Here we report some representative results for such
problems.

\begin{figure}[ht!]
  \begin{center}
    \begin{tabular}{cccccc}
      \widgraph{0.15\textwidth}{\figdir/fig_parrot_student_noise_50_alg_125} &
      \widgraph{0.15\textwidth}{\figdir/fig_parrot_student_noise_50_alg_80} &
      \widgraph{0.15\textwidth}{\figdir/fig_parrot_student_noise_50_alg_60} &
      \widgraph{0.15\textwidth}{\figdir/fig_parrot_student_noise_50_alg_40} &
      \widgraph{0.15\textwidth}{\figdir/fig_parrot_student_noise_50_alg_20} &
      \widgraph{0.15\textwidth}{\figdir/fig_parrot_student_noise_50_alg_0} 
    \end{tabular}
        \begin{tabular}{c}
          \widgraph{0.5\textwidth}{\figdir/fig_mse_parrot_student_noise_25}\\
          \widgraph{0.8\textwidth}{\figdir/fig_psnr_student_noise_parrot_scale_130}
    \end{tabular}
  \end{center}
  \caption{ Performance of wild refitting for MSE estimation for
    plug-and-play denoising with Student-$t$ noise with $d = 6$
    degrees of freedom.
    \protect \input{denoise_caption.tex} Wild refitting with $\rho = 1.30$
    yields the best performance in this case.}
  \label{FigStudent}
\end{figure}

\paragraph{Gaussian case:}
Beginning with the (easiest) case of Gaussian noise, we generated a
noisy version of the Sloth image shown in~\Cref{FigGauss} by choosing
$w_i \sim N(0, \sigma^2)$. Using this noisy image (represented as a
vector $y \in \real^\numobs$, with $\numobs = 512 \times 512$), we
then computed the plug-and-play estimate $\fhat_\gamma$ for a range of
smoothing parameters $\gamma$.  In the top row of~\Cref{FigGauss}, we
plot the denoised images $\fhat_\gamma$ for five different choices of
$\gamma$, ranging from a larger value (over-smoothed) over to a
smaller value (under-smoothed).  In panel (a) of~\Cref{FigGauss}, we
plot the true MSE $\|\fhat_\gamma - \fstar\|_\numobs^2$ versus the
inverse log smoothing parameter $\log(1/\gamma)$, so that the left
side (respectively right side) corresponds to under-smoothing
(respectively over-smoothing).  We also plot the wild refitting
estimates of the MSE for three different choices of the noise scale:
$\rho \in \{1.0, 1.15, 1.30 \}$.  Consistent with our theory, the wild
refitting estimates are upper bounds on the true MSE once $\rho$ is
sufficiently large; in this example, setting $\rho = 1.15$ is
sufficient.

Panel (b) in~\Cref{FigGauss} compares the MSE obtained by the oracle
choice $\gamma^*$ to the MSE obtained by the wild refitting choice
$\gamwild$; in each case, we plot these MSEs versus the true standard
deviation $\sigma$ of the additive Gaussian noise variables ($w \sim
N(0, \sigma^2)$).  The left-middle-right panels correspond to the wild
noise scales $\rho \in \{1.0, 1.15, 1.30 \}$.  In this particular
case, we see that with the choice $\rho = 1.15$, the MSE obtained from
the wild choice $\gamwild$ is \emph{identical} to the oracle MSE.

\paragraph{Student-$t$ noise:}  We then moved onto a more
challenging case of heavy-tailed noise, in particular choosing the
$w_i$-variables to have a Student-$t$ distribution with $d = 6$
degrees of freedom.  Focusing on the Parrot image shown
in~\Cref{FigStudent}, we again computed the plug-and-play estimate
$\fhat_\gamma$ for a range of smoothing parameters $\gamma$, and we
plot these denoised images in the top row of~\Cref{FigStudent}, for
five different choices of $\gamma$.  Panel (a) of~\Cref{FigStudent}
gives plots of the true MSE $\|\fhat_\gamma - \fstar\|_\numobs^2$ and
wild refitting estimates of the MSE versus the inverse log smoothing
parameter $\log(1/\gamma)$.  We plot the wild refitting estimates of
the MSE for three different choices of the noise scale: $\rho \in
\{1.0, 1.15, 1.30 \}$.  Consistent with our theory, the wild refitting
estimates are upper bounds on the true MSE once $\rho$ is sufficiently
large; in this example, setting $\rho = 1.30$ is adequate.  It is
worth noting that the wild MSE estimate with $\rho = 1.0$ is a severe
under-estimate; in fact, the MSE estimate is negative for certain
values of $\gamma$.

Panel (b) in~\Cref{FigStudent} compares the MSE obtained by the oracle
choice $\gamma^*$ to the MSE obtained by the wild refitting choice
$\gamwild$.  Again, we plot these MSEs versus the standard deviation
$\sigma$ of the underlying noise variables (Student-$t$ in this case).
The left-middle-right panels correspond to the wild noise scales $\rho
\in \{1.0, 1.15, 1.30 \}$; for this Student-$t$ noise,
the choice $\rho = 1.30$ yields the best approximation to the
oracle MSE.

\begin{figure}[h!]
  \begin{center}
    \begin{tabular}{cccccc}
      \widgraph{0.15\textwidth}{\figdir/fig_starfish_half_noise_50_alg_125}&
      \widgraph{0.15\textwidth}{\figdir/fig_starfish_half_noise_50_alg_100}&
      \widgraph{0.15\textwidth}{\figdir/fig_starfish_half_noise_50_alg_80}&
      \widgraph{0.15\textwidth}{\figdir/fig_starfish_half_noise_50_alg_30}&
      \widgraph{0.15\textwidth}{\figdir/fig_starfish_half_noise_50_alg_10}&
      \widgraph{0.15\textwidth}{\figdir/fig_starfish_half_noise_50_alg_0}
    \end{tabular}
        \begin{tabular}{c}
          \widgraph{0.5\textwidth}{\figdir/fig_mse_starfish_half_noise_25}  \\
          \widgraph{0.8 \textwidth}{\figdir/fig_psnr_half_noise_starfish_scale_130}
    \end{tabular}
  \end{center}
  \caption{ Performance of wild refitting for MSE estimation for
    plug-and-play denoising with a heterogeneous form of noise: the
    middle strip has high variance, with the outer strips having very
    low variance.  \protect \input{denoise_caption.tex} Wild refitting
    with $\rho = 1.30$ yields the best performance among these three
    choices, but does not match the oracle performance exactly,
    especially for larger values of $\sigma$.}
  \label{FigHalf}
\end{figure}

\paragraph{Highly heteroskedastic noise:}  As previously
discussed, an important feature of wild refitting is that it allows
for heteroskedastic noise.  Accordingly, for our final set of
experiments, we performed denoising when the contaminating noise
variables $w_i$ were highly heteroskedastic.  As illustrated in the
rightmost image in top row of~\Cref{FigHalf}, we contaminated the
Starfish image with noise that has a very high standard deviation
$\sigma_1$ in a central band, and a much smaller standard deviation
$\sigma_0 \ll \sigma_1$ outside of this central band.  We then applied
the plug-and-play denoiser to compute denoised images $\fhat_\gamma$
across a range of smoothing levels $\gamma$, as shown in the remaining
panels.  Panel (a) of~\Cref{FigHalf} gives plots of the true MSE
$\|\fhat_\gamma - \fstar\|_\numobs^2$ and wild refitting estimates of
the MSE versus the inverse log smoothing parameter $\log(1/\gamma)$.
We plot the wild refitting estimates of the MSE for three different
choices of the noise scale: $\rho \in \{1.0, 1.15, 1.30 \}$.
Consistent with our theory, the wild refitting estimates are upper
bounds on the true MSE once $\rho$ is sufficiently large; in this
example, setting $\rho = 1.30$ is adequate.

Panel (b) in~\Cref{FigHalf} compares the MSE obtained by the oracle
choice $\gamma^*$ to the MSE obtained by the wild refitting choice
$\gamwild$.  We plot these MSEs versus the square root of the average
variance: that is, the quantity $\sigma \defn \sqrt{\frac{1}{\numobs}
  \sum_{i=1} \sigma^2(x_i)}$, where $\sigma^2(x_i)$ is the variance of
the noise at image position $x_i$.  The left-middle-right panels
correspond to the wild noise scales $\rho \in \{1.0, 1.15, 1.30 \}$;
Among these three choices, setting $\rho = 1.30$ yields the best
approximation to the oracle MSE, but we see that it does not exactly
match the oracle in this more challenging case.


\section{Proofs}
\label{SecProofs}

In this section, we turn to the proofs of our main claims,
including~\Cref{ThmWildOptBound,ThmRhatError} in
in~\Cref{SecThmWildOptBound,SecThmRhatError}, respectively.


\subsection{Proof of~\Cref{ThmWildOptBound}}
\label{SecThmWildOptBound}

Our proof involves relating the optimism and its wild variant
via the intermediate quantities
\begin{subequations}
\begin{align}
\OptDagger(\fhat) & = \frac{1}{\numobs} \sum_{i=1}^\numobs w_i
\big(\fhat(x_i) - \fproj(x_i) \big), \quad \mbox{and} \\
  \ZoptRade(r) & \defn \sup_{f \in \MyBall{\fproj}{r}} \Big[ \FULLSUM
    \rade{i} \noise_i \big(f(x_i) - \fproj(x_i) \big) \Big].
\end{align}
\end{subequations}

\noindent The following lemma shows how they can be used to bound the
optimism:
\begin{lemma}
\label{LemOptBound}
For any $\tpar > 0$, we have
\begin{subequations}  
\begin{align}
\label{EqnOptBound}
\Opt(\fhat) & \leq \OptDagger(\fhat) + \empnorm{\fproj - \fstar} \:
\DevTerm \qquad \mbox{with probability at least $1 - e^{-t^2}$.}
\end{align}
Moreover, for any $r \geq \|\fhat - \fproj\|_\numobs$ and any $t > 0$,
we have
\begin{align}
\label{EqnZoptRadeConc}
\max \big \{ \OptDagger(\fhat), \ZoptRade(r) \big \} & \leq
\ExsRade[\ZoptRade(r)] + r \DevTerm.
\end{align}
\end{subequations}
with probability at least $1 - 2 e^{-t^2}$.
\end{lemma}
\noindent See~\Cref{SecLemOptBound} for the proof of this claim. \\

Based on~\Cref{LemOptBound}, we see that for any $r \geq \|\fhat -
\fproj\|_\numobs$, we have the upper bound
\begin{align}
\label{EqnFizzy}  
  \Opt(\fhat) & \leq \ExsRade[\ZoptRade(r)] + \big \{ r +
  \empnorm{\fproj - \fstar} \big \} \; \DevTerm
\end{align}
\mbox{with probability at least $1 - 3 e^{-t^2}$.}  Our next step is
to relate the expectation $\ExsRade[\ZoptRade(r)]$ to the wild
optimism $\OptWild(\fhat)$.  In order to do so, we make use of the
intermediate family of random variables
\begin{align}
\label{EqnDefnZwildRade}  
  \ZwildRade(r) & \defn \sup_{f \in \MyBall{\fhat}{r}} \Big[ \FULLSUM
    \rade{i} \noise_i \big(f(x_i) - \fhat(x_i) \big) \Big].
\end{align}

\noindent The following auxiliary result relates the expectations of
$\ZoptRade(r)$ and $\ZwildRade(r)$, and involves the estimation error
$\rhat \defn \empnorm{\fhat - \fproj}$.
\begin{lemma}
\label{LemZopt2Wild}
We have the deterministic bound
\begin{subequations}
\begin{align}
\label{EqnZopt2Wild}
\ExsRade[\ZoptRade(r)] & \leq \ExsRade[\ZwildRade(r + \rhat)]
\end{align}
along with the bound
\begin{align}
\label{EqnZwildRadeConc}
\ExsRade[\ZwildRade(r)] & \leq \ZwildRade(r) +  r \DevTerm \qquad
\mbox{with probability at least $1 - e^{-t^2}$.}
\end{align}
\end{subequations}
\end{lemma}
\noindent See~\Cref{SecLemZopt2Wild} for the proof of this claim.
\\

The two bounds from~\Cref{LemZopt2Wild} imply that
for any $r \geq \rhat$, we have
\begin{align*}
  \ExsRade[\ZoptRade(2 r)] & \leq \ZwildRade(2 r) + (2 r) \;  \DevTerm
\end{align*}
with probability at least $1 - e^{-t^2}$.
Combined with our earlier inequality~\eqref{EqnFizzy}, we have shown that
\begin{align}
\label{EqnPurpleGlasses}  
  \Opt(\fhat) & \leq \ZwildRade(2 r) + \big \{ 3 r + \empnorm{\fproj -
    \fstar} \big \} \; \DevTerm
\end{align}
with probability at least $1 - 4 e^{-t^2}$.  Our final step is to
relate $\ZwildRade(2 r)$ to the wild optimism $\OptWild(\fwild)$ for a
suitably chosen noise scale $\rho$.

\begin{lemma}
  \label{LemFinal}
For any radius $r$, we have 
\begin{align}
\label{EqnFinal}  
\ZwildRade(2 r) & \leq \OptWild(\fwild) + \AppError,
  \end{align}
where $\fwild$ is the wild solution with $\|\fwild - \fhat\|_\numobs =
2r$.
\end{lemma}
\noindent See~\Cref{SecProofLemFinal} for the proof of this claim. \\

Finally, by applying the bound~\eqref{EqnFinal} to our earlier
inequality~\eqref{EqnPurpleGlasses}, we find that
\begin{align*}
  \Opt(\fhat) & \leq \OptWild(\fwild) + \big \{ 3 r + \empnorm{\fproj
    - \fstar} \big \} \; \DevTerm + \AppError,
\end{align*}
with probability at least $1 - 4 e^{-t^2}$, as claimed
in~\Cref{ThmWildOptBound}.


\subsection{Auxiliary lemmas for~\Cref{ThmWildOptBound}}
\label{SecAuxiliaryLemmas}
In this section, we collect together the proofs of the auxiliary
lemmas used in the proof of~\Cref{ThmWildOptBound}.


\subsubsection{Proof of~\Cref{LemOptBound}}
\label{SecLemOptBound}

We split our analysis into two parts, corresponding to the
two claims in the lemma.

\paragraph{Proof of the bound~\eqref{EqnOptBound}:}
Beginning with the definition of the optimism~\eqref{EqnDefnOptimism},
we can add and subtract terms involving $\fproj$, thereby obtaining
\begin{align*}
\Opt(\fhat) = \underbrace{\empinner{\wnoise}{\fhat -
    \fproj}}_{\OptDagger(\fhat)} + \underbrace{\empinner{\wnoise}{\fproj -
    \fstar}}_{\Term}.
\end{align*}
It remains to bound the term $\Term$.  Since variable $\wnoise_i$ has
a distribution symmetric around zero, we can write $\wnoise_i =
\rade{i} |\wnoise_i|$, where $\rade{i} \in \{-1, +1 \}$ is a random
sign (Rademacher variable) independent of $|\wnoise_i|$.  Conditioning
on the absolute values, we can represent $\Term$ as a realization of
the linear (and hence convex) function
\begin{align*}
\big( \rade{1}, \ldots, \rade{\numobs} \big) & \mapsto G(\rade{})
\defn \empinner{ \rade{} |\wnoise|}{\fproj - \fstar} \; = \; \Big \{
\frac{1}{\numobs} \sum_{i=1}^\numobs \rade{i} |\wnoise_i|
\big(\fproj(x_i) - \fstar(x_i) \big) \Big \}.
\end{align*}
Letting $\rade{}$ and $\rade{}'$ be two Rademacher sequences,
applying the Cauchy--Schwarz inequality yields
\begin{align*}
  \big| G(\rade{}) - G(\rade{}') \big| \; = \; \FULLSUM (\rade{i} -
  \rade{i}') \, |\wnoise_i| \big(\fproj(x_i) - \fstar(x_i) \big)
  & \leq \frac{1}{\numobs} \big(\sum_{i=1}^\numobs \wnoise_i^2 \big(
  \fproj(x_i) - \fstar(x_i) \big)^2\big)^{1/2} \; \|\rade{} -
  \rade{}'\|_2 \\
 & \leq \tfrac{\|\wnoise\|_\infty}{\sqrt{\numobs}} \empnorm{\fproj -
    \fstar} \, \|\rade{} - \rade{}'\|_2
\end{align*}
showing that the function $G$ is $L$-Lipschitz with $L \defn
\tfrac{\|\wnoise\|_\infty}{\sqrt{\numobs}} \empnorm{\fproj - \fstar}$.
Since $\Exs_{\rade{}} [G(\rade{})] = 0$, standard concentration
results for convex/Lipschitz functions of Rademacher variables
(cf. Theorem 3.24 in the book~\cite{Wai19}) guarantee that
\begin{align*}
G(\rade{}) & \leq \tfrac{2 \|\wnoise\|_\infty \tpar}{\sqrt{\numobs}}
\; \empnorm{\fproj - \fstar} \qquad \mbox{with probability at least $1
  - e^{-\tpar^2}$.}
\end{align*}

\paragraph{Proof of
the bound~\eqref{EqnZoptRadeConc}:} We condition on the vector
$|\wnoise|$ of absolute values throughout this proof.  Letting
$\rade{} \in \{-1, +1 \}^\numobs$ be an i.i.d. vector of Rademacher
variables, we can view the random variable $\ZoptRade(r)$ as a
function
\begin{align*}
  \rade{} \mapsto H(\rade{}) \defn \ZoptRade(r) \defn
  \sup_{f \in
  \MyBall{\fproj}{r}} \Big[ \FULLSUM \rade{i} |\noise_i| \big(f(x_i) -
  \fproj(x_i) \big) \Big].
\end{align*}
As in our previous proof, we can write $\noise_i = \radetil_i
|\noise_i|$, where $\radetil_i \in \{-1, +1 \}$ is a Rademacher
variable independent of $|\noise_i|$.  Thus, we recognize the random
variable $\Zopt(r)$ as a particular realization of the random variable
defined by $H$.  Consequently, bounding the difference $\Zopt(r) -
\ExsRade[\ZoptRade(r)]$ is equivalent to establishing an upper tail
bound for $H(\rade{})$ in terms of its expectation
$\ExsRade[H(\rade{})]$.  In particular, the claim is equivalent to
showing that $H(\rade{}) \leq \ExsRade[H(\rade{})] + r \: (\DevTerm)$
with probability at least $1 - e^{-\tpar^2}$.

In order to establish this claim, we again make use of concentration
bounds for convex and Lipschitz functions of independent bounded
random variables (cf. Theorem 3.24 in the book~\cite{Wai19}).  As a
supremum of linear functions, the function $H$ is convex.  The claimed
bound follows as long as we can establish the Lipschitz bound
\begin{align}
\label{EqnHclaim}  
  |H(\rade{}) - H(\rade{}')| & \leq L \|\rade{} - \rade{}'\|_2 \qquad
  \mbox{for $L = \tfrac{\|\wnoise\|_\infty}{\sqrt{\numobs}} r$,}
\end{align}
valid for any pair of Rademacher vectors $\rade{}, \rade{}' \in \{-1,
+1 \}^\numobs$.

Let $g$ by any function that achieves the supremum defining
$H(\rade{})$.  We can then write
\begin{align*}
  H(\rade{}) - H(\rade{}') & \leq \Big[ \FULLSUM \rade{i} |\noise_i|
    \big(g(x_i) - \fproj(x_i) \big) \Big] - \Big[ \FULLSUM \rade{i}'
    |\noise_i| \big(g(x_i) - \fproj(x_i) \big) \Big] \\
  & = \FULLSUM \big(\rade{i} - \rade{i}' \big) |\noise_i| \big(g(x_i)
  - \fproj(x_i) \big).
\end{align*}
Applying the Cauchy--Schwarz inequality yields
\begin{align*}
  H(\rade{}) - H(\rade{}') \; \leq \; \frac{1}{\numobs} \big \{
  \sum_{i=1}^\numobs \noise_i^2 \big(g(x_i) - \fproj(x_i) \big)^2 \big
  \}^{1/2} \; \big( \|\rade{} - \rade{}'\|_2\big) & \leq
  \frac{\|\wnoise\|_\infty}{\sqrt{\numobs}} \empnorm{g - \fproj}
  \|\rade{} - \rade{}'\|_2 \\ & \stackrel{(\star)}{\leq}
  \frac{\|\wnoise\|_\infty}{\sqrt{\numobs}} \, r \, \|\rade{} -
  \rade{}'\|_2.
\end{align*}
where the final bound $(\star)$ follows since $\empnorm{g - \fproj}
\leq r$.  The same argument applies with the roles of $\rade{}$ and
$\rade{}'$ reversed, from which the claimed Lipschitz
bound~\eqref{EqnHclaim} follows.


\subsubsection{Proof of~\Cref{LemZopt2Wild}}
\label{SecLemZopt2Wild} 

We prove each of the two claims in turn.

\paragraph{Proof of the bound~\eqref{EqnZopt2Wild}:}
Let $g \in \MyBall{\fproj}{r}$ be any function that achieves the
supremum defining $\ZoptRade(r)$.  In terms of this function, we have
the decomposition $\ZoptRade(r) = \FULLSUM \rade{i} \noise_i
\big(g(x_i) - \fproj(x_i) \big) = \Term_1 + \Term_2$, where
\begin{align*}
\Term_1 & \defn \FULLSUM \rade{i} \noise_i \big(g(x_i) - \fhat(x_i)
\big), \quad \mbox{and} \quad \Term_2 \defn \FULLSUM \rade{i} \noise_i
\big(\fhat(x_i) - \fproj(x_i) \big).
\end{align*}
By the triangle inequality, we have
\begin{align*}
\empnorm{g - \fhat} & \leq \empnorm{g - \fproj} + \empnorm{\fproj -
  \fhat} \; \leq \; 2 r.
\end{align*}
where the final step follows since $\fhat$ and $g$ belong the ball
$\MyBall{\fproj}{r}$. Consequently, we have shown that
\begin{align*}
\Term_1 & \leq \sup_{f \in \MyBall{\fhat}{2 r}} \FULLSUM \rade{i}
\noise_i \big(g(x_i) - \fhat(x_i) \big), \quad \mbox{and hence} \quad
\ExsRade[\Term_1] \leq \ExsRade[ \ZwildRade(2 r)].
\end{align*}
As for the second term, we have
\begin{align*}
\ExsRade[\Term_2] & = \ExsRade \Big[ \FULLSUM \rade{i} \noise_i
  \big(\fhat(x_i) - \fproj(x_i) \big) \Big] \; = \; 0,
\end{align*}
using the fact that $\rade{i}$ is a Rademacher variable independent of
the original noise vector $\wnoise_i$, and hence $\fhat$ as well.
Putting together the pieces, we have shown that
\begin{align*}
\ExsRade[\ZoptRade(r)] & \leq \ExsRade[\Term_1] \; = \;
\ExsRade[\ZwildRade(2 r)],
\end{align*}
thus establishing the claim~\eqref{EqnZopt2Wild}
of~\Cref{LemZopt2Wild}.

\paragraph{Proof of the bound~\eqref{EqnZwildRadeConc}:}
The random variable $\ZwildRade(r)$ has a structure analogous to
$\ZoptRade(r)$, so that this claim follows by an argument entirely
analogous to that used to prove the claim~\eqref{EqnZoptRadeConc}
from~\Cref{LemOptBound}.


\subsubsection{Proof of~\Cref{LemFinal}}
\label{SecProofLemFinal}
By definition of the wild noise $\wtil_i = y_i - \ftil(x_i)$, we have
\begin{align*}
\rade{i} \wnoise_i & = \rade{i} \wtil_i + \rade{i} (\ftil(x_i) -
\fstar(x_i)).
\end{align*}
Consequently, from the definition of  $\ZwildRade$ combined with
the triangle inequality, we can write
\begin{align}
  \ZwildRade(2r) & =
  \sup_{f \in \MyBall{\fhat}{2r}} \Big[ \FULLSUM
    \rade{i} \noise_i \big(f(x_i) - \fhat(x_i) \big) \Big]  \notag \\
  & \leq \sup_{f \in \MyBall{\fhat}{2 r}} \Big[ \FULLSUM \rade{i}
    \wtil_i \big(f(x_i) - \fhat(x_i) \big) \Big] + \sup_{f \in
    \MyBall{\fhat}{2r}} \Big[ \FULLSUM \rade{i} \big (\ftil(x_i) -
    \fstar(x_i) \big) \big(f(x_i) - \fhat(x_i) \big) \Big] \notag \\
\label{EqnRavenBound}  
  & = \Zwild(2 r) + \AppError.
\end{align}
Finally, by~\Cref{LemOpt2Complex}, we have the equivalence $\Zwild(2
r) = \OptWild(\fwild)$ for the wild solution $\fwild$ with noise scale
$\rho$ chosen to ensure that $\|\fwild - \fhat\|_\numobs = 2 \rho$.


\subsection{Proof of~\Cref{ThmRhatError}}
\label{SecThmRhatError}

We now turn to the proof of~\Cref{ThmRhatError}, which provides upper
bounds on the estimation error $\rhat \defn \|\fhat -
\fproj\|_\numobs$.


\subsubsection{Proof of the bound~\eqref{EqnRhatError}}

Our proof of this claim requires two auxiliary results, which we begin
by stating.

\begin{lemma}
  \label{LemRhatBasic}
Given a procedure $\Meth$ that is firmly non-expansive~\eqref{EqnFirm}
around $\fstar$, the error \mbox{$\rhat \defn \empnorm{\fhat -
    \fproj}$} satisfies the bound
\begin{align}
\label{EqnRhatBasic}
\rhat^2 \leq \Zopt(\rhat).
\end{align}
\end{lemma}
\noindent See~\Cref{SecLemRhatBasic} for the proof. \\

\noindent We also require the following generalization of the
bound~\eqref{EqnZoptRadeConc} from~\Cref{LemOptBound}.  In
particular, it allows us to bound $\Zopt(r)$ for a random radius (such
as $\rhat$).
\begin{lemma}
\label{LemZconcRandom}
For any scalar $s \geq 3$, we have
\begin{align}
\label{EqnColdUpper}  
\Zopt(r) & \leq \ExsRade \big[\ZoptRade \big([\oneplus] \; r \big)
  \big] + \tfrac{4 \|\noise\|_\infty}{s} \: r^2
\end{align}
uniformly for all $r \geq \frac{s^2}{\sqrt{\numobs}}$ with
probability at least $1 - e^{-s^2}$.
\end{lemma}
\noindent See~\Cref{SecLemZconcRandom} for the proof. \\

Equipped with these lemmas, let us now complete the proof of the
theorem.  Fix some $s \geq 3$.  We either have $\rhat \leq
s^2/\sqrt{\numobs}$, or we may assume that $\rhat >
s^2/\sqrt{\numobs}$.  The remainder of our proof assumes that the
latter inequality holds.

We have
\begin{align*}
  \rhat^2 & \stackrel{(i)}{\leq} \Zopt(\rhat) \; \stackrel{(ii)}{\leq}
  \; \ExsRade \big[\ZoptRade \big([\oneplus] \; \rhat \big) \big] +
  \tfrac{4 \|\noise\|_\infty}{s} \: \rhat^2,
\end{align*}
where step (i) follows from~\Cref{LemRhatBasic}; and inequality (ii)
holds with probability at least $1 - e^{-s^2}$, based
on~\Cref{LemZconcRandom}.  By applying the bound~\eqref{EqnZopt2Wild}
from~\Cref{LemZopt2Wild} with $r = \rhat$, we find that
\begin{subequations}
\begin{align}
  \label{EqnAustinOne}
\rhat^2 & \leq \ExsRade \big[\ZwildRade \big([\twoplus] \; \rhat \big)
  \big] + \tfrac{4 \|\noise\|_\infty}{s} \: \rhat^2.
\end{align}
Next we apply the bound~\eqref{EqnZwildRadeConc}
from~\Cref{LemOptBound} with the choice $t = \rhat \sqrt{\numobs}/s$,
thereby obtaining\footnote{To be clear, while $\rhat$ is a random
radius, this randomness is independent of the Rademacher randomness
that defines $\ZwildRade$, so that we do not to use a uniform radius
result.}
\begin{align}
\label{EqnAustinTwo}  
\ExsRade[\ZwildRade \big ([\twoplus] \rhat \big)] & \leq \ZwildRade
\big([\twoplus] \; \rhat \big) + \tfrac{2 \|\noise\|_\infty}{s} \:
\rhat^2,
\end{align}
\end{subequations}
with probability at least $1 - e^{-t^2}$.  With our choice $t = \rhat
\sqrt{\numobs}/s$, we have $t^2 = \rhat^2 \numobs/s^2 \geq s^2$ since
we have assumed that $\rhat \geq s^2/\sqrt{\numobs}$ in this portion
of the argument; as consequence, the bound~\eqref{EqnAustinTwo} holds
with probability at least $1 - e^{-s^2}$.

By combining inequalities~\eqref{EqnAustinOne}
and~\eqref{EqnAustinTwo}, with probability at least $1 - 2 e^{-s^2}$,
we have
\begin{align*}
  \rhat^2 & \leq \ZwildRade \big([\twoplus] \; \rhat \big) + \tfrac{4
    \|\noise\|_\infty}{s} \: \rhat^2 \; \stackrel{(\star)}{\leq} \;
  \Zwild \big([\twoplus] \; \rhat \big) + \tfrac{4
    \|\noise\|_\infty}{s} \: \rhat^2 + \AppError,
\end{align*}
where step $(\star)$ follows inequality~\eqref{EqnRavenBound} from the
proof of~\Cref{LemFinal} in~\Cref{SecProofLemFinal}.  (In particular,
we use the bound~\eqref{EqnRavenBound} with the choice $ r = 1/2 \big(
[\twoplus] \; \rhat)$.)

Modulo the replacement of $s$ by $t$, we have thus established the
claim~\eqref{EqnRhatError} from~\Cref{ThmRhatError}.  (The additional
term $\tpar^2/\numobs$ arises from taking into account the possibility
that $\rhat \leq \tpar/\sqrt{\numobs}$ in our argument.)

\subsubsection{Proof of the bound~\eqref{EqnWildRefit}}
\label{SecCorWildRefit}
In order to prove the bound~\eqref{EqnWildRefit}, we require the
following auxiliary result.
\begin{lemma}
\label{LemZwildConcave}  
For any convex function class $\Fclass$, the function $u \mapsto
\Zwild(u)$ is concave on the interval $[0, \infty]$, and hence we have
the non-increasing property
  \begin{align}
    \label{EqnZwildConcave}
    \frac{\Zwild(s)}{s} \leq \frac{\Zwild(t)}{t} \qquad \mbox{for any
      $s \geq t > 0$.}
  \end{align}
\end{lemma}
\noindent See~\Cref{SecLemZwildConcave} for the proof.

Let us now prove the stated bound~\eqref{EqnWildRefit}.  Either we
have $\rhat \leq \rwild$, or $\rhat > \rwild$.  In the latter case, we
can write
\begin{align*}
\Zwild([\twoplust] \rhat) \; = \; [\twoplust] \rhat
\tfrac{\Zwild([\twoplust] \rhat)}{[\twoplust] \rhat} & \; \leq \;
      [\twoplust] \rhat \tfrac{\Zwild([\twoplust] \rwild)}{[\twoplust]
        \rwild} \\
& \; = \; \rhat \tfrac{\Zwild([\twoplust] \rwild)}{ \rwild},
\end{align*}
where the inequality follows from equation~\eqref{EqnZwildConcave}
in~\Cref{LemZwildConcave}.  Combining with
inequality~\eqref{EqnRhatError} from~\Cref{ThmRhatError} yields the
claim~\eqref{EqnWildRefit}.

\subsection{Auxiliary lemmas for~\Cref{ThmRhatError}}

In this section, we collect together proofs of the auxiliary
lemmas involved in~\Cref{ThmRhatError}.

\subsubsection{Proof of~\Cref{LemRhatBasic}}
\label{SecLemRhatBasic}

To prove this claim, we first observe that by definition $\fhat =
\Meth(\fstar + \wnoise)$ and $\fproj = \Meth(\fstar)$.  Since the
estimator $\Meth$ is firmly non-expansive around $\fstar$, we have
\begin{align*}
\rhat^2 = \empnorm{\fhat - \fproj}^2 = \empnorm{\Meth(\fstar +
  \wnoise) - \Meth(\fstar)}^2 & \leq \empinner{\wnoise}{\fhat -
  \fproj} \\
& = \FULLSUM \noise_i \big(\fhat(x_i) - \fproj(x_i) \big) \\
& \stackrel{(\dagger)}{\leq} \Zopt(\rhat),
\end{align*}
where inequality $(\dagger)$ follows from the definition of $\Zopt$.

\subsubsection{Proof of~\Cref{LemZconcRandom}}
\label{SecLemZconcRandom}

For an arbitrary $s > 0$, we apply the bound~\eqref{EqnZoptRadeConc}
with $t \defn \numobs r/s$ to obtain
\begin{align}
\label{EqnMammoth}
\Prob\big[\ZnewHat(r) \geq \ExsRade[\ZoptRade(r)] + \tfrac{2
    \|\noise\|_\infty}{s} \; r^2 \big] & \; \stackrel{(i)}{\leq} \;
e^{-\tfrac{r^2}{s^2} \numobs} \; \stackrel{(ii)}{\leq} \; e^{-s^2}.
\end{align}
where the final inequality (ii) follows as long as $r \geq
s^2/\sqrt{\numobs}$.

Now let $\Event$ be the event the bound~\eqref{EqnColdUpper} is
violated for some $r \geq \frac{s^2}{\sqrt{\numobs}}$.  Our strategy
is to exploit the tail bound~\eqref{EqnMammoth}(i) to prove that
$\Prob[\Event] \leq e^{-s^2}$.  In order to do so, we first define the
increasing sequence
\begin{align*}
q_0 \defn \tfrac{s^2}{\sqrt{\numobs}} \quad \mbox{and} \quad q_m \defn
(\oneplus)^m q_0 \qquad \mbox{for $m = 1, 2, 3 \ldots$,}
\end{align*}
and then decompose the event of interest as $\Event = \cup_{m
  =0}^\infty \Event_m$, where
\begin{align*}
  \Event_m \defn \Big \{ \exists r \in [q_m, q_{m+1}) \quad \mbox{such
      that the bound~\eqref{EqnColdUpper} is violated} \Big \}.
\end{align*}
We claim that it suffices to show that
\begin{align}
\label{EqnProvidence}  
\Prob[\Event_m] & \leq e^{-\frac{\numobs}{s^2} q_{m+1}^2} \qquad
\mbox{for each $m = 0, 1, 2, \ldots$.}
\end{align}
Taking this inequality as given, we can then apply the union bound to
obtain
\begin{align*}
\Prob[\Event] \leq \sum_{m = 0}^\infty \Prob[\Event_m] \;
\stackrel{(i)}{\leq} \; \sum_{m=0}^\infty e^{-\frac{\numobs}{s^2}
  q_{m+1}^2} & \stackrel{(ii)}{\leq} e^{-s^2} \sum_{m=0}^\infty e^{- 2
  (m + 1) s} \; \stackrel{(iii)}{\leq} \; e^{-s^2},
\end{align*}
where step (i) follows from the bound~\eqref{EqnProvidence}.  In order
to verify inequality (ii), we observe that
\begin{align*}
 \frac{\numobs}{s^2} q_{m+1}^2 \; = \; \frac{\numobs}{s^2} \, \big(
 \oneplus \big)^{2 (m+1)} q_0^2 & \geq \frac{\numobs}{s^2} q_0^2 \big \{ 1 + 2 (
 m + 1) (1/s) \big \} \\
 & = s^2 + 2 (m + 1) s,
\end{align*}
using the fact that $(1 + (1/s))^{2 (m+1)} \geq 1 + 2 (m+1) (1/s)$,
and the equality $\frac{\numobs}{s^2} q_0^2 = s^2$.  Finally, step
(iii) follows since $\sum_{m=0}^\infty e^{- 2 (m + 1) s} \leq
\sum_{m=1}^\infty (1/2)^m = 1$, since the assumption that $s \geq 3$
implies that $e^{-2 s} \leq \tfrac{1}{2}$.

\paragraph{Proof of the bound~\eqref{EqnProvidence}:}

So as to reduce clutter, introduce the shorthand \mbox{$\Ztrue(r)
  \defn \ExsRade[\ZoptRade(r)]$.}  If the event $\Event_m$ holds, then
we are guaranteed to have some $r \in [q_m, q_{m+1}]$ such that
\begin{align*}
\ZnewHat(q_{m+1}) \stackrel{(i)}{\geq} \ZnewHat(r) &
\stackrel{(ii)}{\geq} \Ztrue( [\oneplus] r) + \tfrac{4
  \|\noise\|_\infty}{s} r^2 \\
& \stackrel{(iii)}{\geq} \Ztrue(q_{m+1}) + \tfrac{4
  \|\noise\|_\infty}{s} q_m^2 \\
& \stackrel{(iv)}{\geq} \Ztrue(q_{m+1}) + \tfrac{2
  \|\noise\|_\infty}{s} q_{m+1}^2,
\end{align*}
where step (i) follows since $q_{m+1} \geq r$ and the function $u
\mapsto \Zopt(u)$ is non-decreasing in $u$; inequality (ii) follows by
definition of the event $\Event_m$; inequality (iii) holds since $r
\geq q_m$ and $[\oneplus] r \geq [\oneplus] q_m = q_{m+1}$; and step
(iv) follows since
\begin{align*}
\tfrac{q_{m}^2}{q_{m+1}^2} & = \frac{1}{[\oneplus]^2} \geq \frac{1}{(1
  + \frac{1}{3})^2} \; = \; \frac{9}{16} \; \geq \; \frac{1}{2}
\end{align*}
using the assumption $s \geq 3$.

From this string of inequalities, we can conclude that
\begin{align*}
  \Prob[\Event_m] & \leq \Prob \Big[ \ZnewHat(q_{m+1}) \geq
    \Ztrue(q_{m+1}) + \tfrac{2 \|\noise\|_\infty}{s} q_{m+1}^2 \Big]
  \; \stackrel{(v)}{\leq} \; e^{-\frac{\numobs}{s^2} q_{m+1}^2},
\end{align*}
where inequality (v) follows by applying the
bound~\eqref{EqnMammoth}(i) with $r = q_{m+1}$.  This completes the
proof of the auxiliary claim~\eqref{EqnProvidence}, and hence the
overall proof.


\subsubsection{Proof of~\Cref{LemZwildConcave}}
\label{SecLemZwildConcave}
Since $\Zwild(0) = 0$, the inequality~\eqref{EqnZwildConcave} is
equivalent to
\begin{align*}
  \frac{\Zwild(s) - \Zwild(0)}{s} & \leq \frac{\Zwild(t) -
    \Zwild(0)}{t} \qquad \mbox{for all $s \geq t > 0$.}
\end{align*}
It is a standard fact from convex analysis~\cite{Roc70} that any
concave function has this property.

Thus, it remains to show that $\Zwild$ is a concave function:
more precisely, we will show that any scalars $s, t \geq 0$
and $\alpha \in [0,1]$, we have
\begin{align*}
  \alpha \Zwild(s) + (1 - \alpha) \Zwild(t) & \leq \Zwild(r) \qquad
  \mbox{where $r \defn \alpha s + (1 - \alpha) t $.}
\end{align*}
Let $f_s$ and $f_t$ be functions achieving the suprema that define
$\Zwild(s)$ and $\Zwild(t)$, respectively, and introduce the shorthand
$f_r = \alpha f_s + (1-\alpha) f_t$.  By the assumed convexity of
$\Fclass$, we have $f_r \in \Fclass$, and moreover, by the triangle
inequality, we have
\begin{align*}
\empnorm{f_r- \fhat} & \leq \alpha \empnorm{f_s - \fhat} + (1- \alpha)
\empnorm{f_t - \fhat} \; \leq \; \alpha s + (1 - \alpha) t \; = \; r.
\end{align*}
Consequently, the function $f_\alpha$ is feasible for the supremum
defining $\Zwild(r)$, so that we can
write
\begin{align*}
\alpha \Zwild(s) + (1 - \alpha) \Zwild(t) & = \FULLSUM \rade{i}
\wtil_i \big( \alpha f_s(x_i) + (1 - \alpha) f_t(x_i) - \fhat(x_i)
\big) \\
& \leq \sup_{f \in \MyBall{r}{\fhat}} \FULLSUM \rade{i} \wtil_i
\big(f(x_i) - \fhat(x_i) \big) \; = \; \Zwild(r),
\end{align*}
as claimed.

\section{Discussion}
\label{SecDiscuss}

In this paper, we have introduced and analyzed the wild refitting
procedure for computing upper bounds on the mean-squared prediction
error of a general black box procedure for regression.  Our procedure
exploits Rademacher symmetrization of the residuals, as in one variant
of the wild bootstrap, hence its name.  In contrast to bootstrap
analysis, our results are non-asymptotic in nature, and require only a
single black box query.  We provided some theory that guides the use
of wild refitting, with the choice of the wild noise scale needed to
ensure upper bounds on the MSE.

The analysis of this paper focused exclusively on the ``fixed design''
setting, in which the covariates are viewed as fixed.  It would be
interesting to extend our analysis to random design as well.
Moreover, while this paper focused on procedures based on quadratic
loss, the basic ideas of wild refitting can be developed more
generally for other types of black box $M$-estimators; we leave this
direction for future work.

\subsubsection*{Acknowledgements}
This work was partially supported by the Cecil H. Green Chair, ONR
grant N00014-21-1-2842 from the Office of Naval Research, and NSF
DMS-2311072 from the National Science Foundation.  It was inspired by
interactions with Peter B\"{u}hlmann at the Oberwolfach Research
Institute, along with repeated in-person demonstrations of his dancing
skills.

\section{Proof of~\Cref{LemOpt2Complex}}
\label{SecLemOpt2Complex}

In this lemma statement, the penalty function $\Pen$ is the indicator
for membership in the convex set $\Convex$.  With a slight abuse of
notation, we write $f \in \Convex$ as a shorthand for $(f(x_1),
\ldots, f(x_\numobs)) \in \Convex$.  By definition, the wild estimate
$\fwild$ is a constrained minimizer of the objective
\begin{align*}
\arg \min_{f \in \Convex} \Big \{ \frac{1}{2 \numobs}
\sum_{i=1}^\numobs \big(\ywild_i - f(x_i) \big)^2 \Big \} \; = \; \arg
\min_{f \in \Convex} \Big \{ \frac{1}{2} \|f - \fhat\|_\numobs^2 -
\rho \frac{1}{\numobs} \sum_{i=1}^\numobs \rade{i} \wtil_i (f(x_i) -
\fhat(x_i)) \Big \}.
\end{align*}
Equivalently, for each radius $r \geq 0$, define the shell $\Convex(r)
= \{f \in \Convex \mid \|f - \fhat\|_\numobs = r \}$.  We then
write
\begin{align*}
\min_{f \in \Convex} \Big \{ \frac{1}{2} \|f - \fhat\|_\numobs^2 -
\rho \frac{1}{\numobs} \sum_{i=1}^\numobs \rade{i} \wtil_i (f(x_i) -
\fhat(x_i)) \Big \} & = \min_{r \geq 0} \min_{f \in \Convex(r)} \Big
\{ \frac{r^2}{2} - \rho \frac{1}{\numobs} \sum_{i=1}^\numobs \rade{i}
\wtil_i (f(x_i) - \fhat(x_i)) \Big \} \\
& = \min_{r \geq 0} \Big \{ \frac{r^2}{2} - \rho \Zwild(r) \Big \},
\end{align*}
using the definition of the wild complexity.

The left-hand side is minimized at $\fwild$ whereas the right-hand
side is minimized at $\rwild = \|\fwild - \fhat\|_\numobs$.
Evaluating the left-hand side at $\fwild$ yields
\begin{align*}
\frac{1}{2} \|\fwild - \fhat\|_\numobs^2 - \rho \frac{1}{\numobs}
\sum_{i=1}^\numobs \rade{i} \wtil_i (\fwild(x_i) - \fhat(x_i)) \Big \}
& \; = \; \frac{(\rwild)^2}{2} - \rho \OptWild(\fwild)
\end{align*}
Evaluating the right-hand side at $\rwild$ yields
\begin{align*}
  \frac{(\rwild)^2}{2} - \rho \Zwild(\rwild) = \frac{(\rwild)^2}{2} -
  \Zwild \big( \|\fwild - \fhat\|_\numobs \big),
\end{align*}
and by comparing these expressions, we see that $\Zwild(\rwild) =
\OptWild(\fwild)$ for any $\rho > 0$, as claimed.



\bibliographystyle{plain}
\bibliography{mjwain_super,martin_papers,new_papers}


\end{document}